\documentclass[journal,twoside,web]{ieeecolor}
\usepackage{generic}
\usepackage{cite}
\usepackage{amsmath,amssymb,amsfonts}
\usepackage{algorithmic}
\usepackage{graphicx}
\usepackage{algorithm,algorithmic}
\usepackage{hyperref}
\hypersetup{hidelinks=true}
\usepackage{textcomp}

\usepackage{xcolor}
\allowdisplaybreaks
\usepackage{multirow}
\usepackage{multicol}

\usepackage{booktabs,tabularx,xcolor}
\usepackage{subcaption}
\usepackage{float}
\usepackage{bm}
\usepackage{bbding}
\usepackage{pifont}
\newcommand{\cmark}{\ding{51}}%
\newcommand{\xmark}{\ding{55}}%

\definecolor{best}{HTML}{c00000}   
\definecolor{second}{HTML}{0066cc} 

\definecolor{color1}{rgb}{0.8,0.1,0.9}  
\definecolor{color2}{rgb}{0.9,0.4,0.1}  
\definecolor{color3}{rgb}{0.1,0.6,0.15} 
\definecolor{color4}{rgb}{0.0,0.6,0.9}  
\definecolor{colorR2}{rgb}{0.85,0.0,0.0}

\def\BibTeX{{\rm B\kern-.05em{\sc i\kern-.025em b}\kern-.08em
    T\kern-.1667em\lower.7ex\hbox{E}\kern-.125emX}}
\begin{document}
\title{Deep Auxiliary Learning via Bilevel Feature Optimization for Label-Efficient Medical Image Segmentation}
\author{Fuping Wu, Xi Jia, Le Zhang, Jiahua Li, Rima Mustafa, Thomas E. Nichols, Bart\l omiej W. Papie\.z
\thanks{This work was supported by the funding from Novo Nordisk.}
\thanks{Fuping Wu (Email: f.wu1@imperial.ac.uk) was with Big Data Institute, Nuffield Department of Population Health, University of Oxford. He is now with Imperial College London.}
\thanks{Xi Jia (Email: xi.jia@psych.ox.ac.uk), Jiahua Li and Rima Mustafa are with the Nuffield Department of Population Health, University of Oxford. Le Zhang is with the University of Birmingham.}
\thanks{Thomas  E. Nichols is with Big Data Institute, Nuffield Department of Population Health, and Centre for Integrative Neuroimaging, FMRIB, Nuffield Department of Clinical Neurosciences, University of Oxford}
\thanks{Bart\l omiej W. Papie\.z is with Big Data Institute, Nuffield Department of Population Health,  University of Oxford.}
}

\maketitle

\begin{abstract}
Auxiliary learning (AL), which can improve the performance of a primary task by leveraging related auxiliary tasks, is commonly regarded as a special case of multi-task learning.
Existing AL approaches typically require manually annotated auxiliary labels, increasing annotation costs, particularly in medical imaging where expert labeling is expensive.
In this work, we investigate whether segmentation performance can be improved using auxiliary tasks whose labels are automatically derived from existing segmentation annotations, requiring no additional manual labeling.
To address this, we propose a novel auxiliary learning framework, formulated as a bilevel optimization problem. Our method enables effective knowledge transfer between tasks via a feature communication module, enhancing the primary segmentation task. 
We evaluate the method on three medical image segmentation benchmarks under limited-label settings. 
The results demonstrate consistent improvements over state-of-the-art methods, achieving superior label efficiency. The code is publicly available at \url{https://github.com/FupingWu90/ALBO}.
\end{abstract}


\section{Introduction}
\label{sec:intro}
\IEEEPARstart{D}{eep} learning has achieved remarkable success in medical image segmentation. However, its performance is highly dependent on large-scale annotated datasets, the creation of which is time-consuming and requires domain expertise \cite{isensee2021nnu,ma2024segment}. This limitation has prompted research into label-efficient learning strategies.

Auxiliary learning (AL), a specialized form of multi-task learning (MTL), has shown strong potential to improve the performance of a primary task under limited supervision by exploiting related auxiliary tasks, and therefore is well-suited to weakly supervised settings such as semi-supervised learning \cite{luo2021semi}.
Unlike MTL, which treats all tasks equally, AL prioritizes a designated main task through auxiliary supervision.
Despite this encouraging progress, especially on natural image benchmarks \cite{shamsian2023auxiliary}, two key challenges remain unresolved for segmentation.

The first is the design of label-efficient auxiliary tasks. Existing methods typically rely on either (i) additional annotated auxiliary tasks such as depth estimation~\cite{standley2020tasks}, or (ii) self-supervised objectives that benefit from large corpora of unlabeled images~\cite{chen2020big, hwang2020self}.
Although effective, these strategies often require additional annotation, auxiliary modalities, or extensive unlabeled datasets, which may be difficult to obtain in medical imaging because of annotation cost and privacy constraints \cite{altaf2019going}.
In this work, we address a fundamental question: \textit{Through AL, can models improve the label efficiency of the target (main) task using auxiliary tasks, whose labels are derived solely from the existing data? And if so, how?} 
Previous studies have employed handcrafted or algorithmically generated labels, such as object boundaries, to regularize segmentation outputs \cite{chen2017dcan,wang2022active,wang2022boundary}. 
However, these approaches are evaluated only when abundant labeled images are available, and their robustness under limited-label settings is not discussed.
Notably, our setting assumes only the existing annotations of the main task: the auxiliary labels are generated from these annotations, and the training set is not expanded with additional unlabeled images. This is the key difference from semi-supervised learning (SSL), which improves label efficiency by exploiting a pool of unlabeled data; the two strategies are therefore complementary, as discussed in Sec.~\ref{sec:albo_ssl}.

\begin{figure*}[tb]
    \centering
    \includegraphics[width=1.0\linewidth]{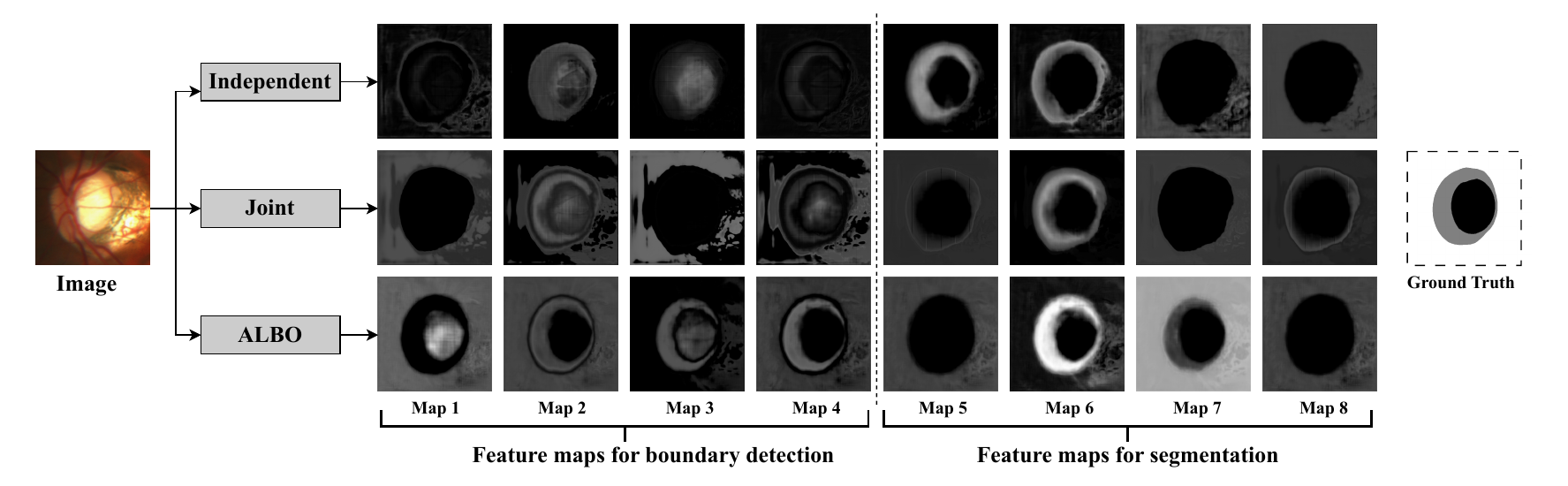}
    \caption{Visualization of feature maps from the encoder of Independent, Joint, and ALBO for a random REFUGE sample.  
    }
    \label{fig:refuge_feature}
\end{figure*}

The second challenge is to optimize the AL models such that the auxiliary task consistently benefits, rather than interferes with, the primary segmentation task.
Simultaneously learning multiple tasks can result in \textit{negative transfer}, where the performance of the main task deteriorates \cite{zhang2022survey}.
Existing approaches primarily attempt to mitigate this either by dynamically adjusting task importance during training \cite{shi2020auxiliary}, or modifying gradients to reduce conflict between tasks \cite{liu2021conflict,yu2020gradient}.
However, the underlying causes of negative transfer remain underexplored, particularly in medical imaging settings where tasks are often highly imbalanced in difficulty and representation \cite{jiang2024forkmerge}.
As illustrated in Fig. \ref{fig:refuge_feature}, without an appropriate optimization technique, the features learned for boundary detection by methods such as Joint and Independent can be inconsistent with this auxiliary task.

To address these challenges, we propose a new auxiliary learning approach for medical image segmentation in which auxiliary labels are automatically derived from the main-task data, thereby eliminating the need for additional manual labeling or dataset expansion. The intuition is that models can benefit from learning auxiliary tasks that capture complementary structural cues (e.g., boundaries or foreground masks) to improve feature representations for the main segmentation task.
Building on this idea, we propose ALBO: \textbf{a}uxiliary \textbf{l}earning via \textbf{b}ilevel \textbf{o}ptimization.
Our main contributions are summarized as follows: 
\begin{itemize}
    \item We propose ALBO, an auxiliary learning framework for medical image segmentation that exploits auxiliary tasks whose labels are automatically derived from existing segmentation annotations. A bidirectional communication module enables feature exchange between the primary and auxiliary branches.

    \item We formulate AL as a bilevel optimization problem and employ a finite-difference approximation to make the optimization tractable during training while enhancing representation learning.

    \item The proposed method is evaluated on two 2D and one 3D benchmark datasets and three types of backbones (convolutional U-Net, ViT-based DINOv3, and Mamba architecture), demonstrating superior label efficiency. 

\end{itemize}

\section{Related Works}
\subsection{Auxiliary learning (AL)}\label{sec:related_al}
AL enhances the performance of a main task by leveraging knowledge from related auxiliary tasks \cite{chen2022module}.
Prior AL research has primarily focused on two fronts: auxiliary task selection and optimization strategies.
Several works have investigated how to identify effective auxiliary tasks by modeling task relationships, in which a common strategy is to group tasks based on their positive transfer potential \cite{fifty2021efficiently,song2022efficient}. 
Another line of research focuses on improving the optimization of AL models.
 In GCS \cite{du2018adapting}, gradient cosine similarity is employed to assess task-relatedness and to determine whether an auxiliary loss is beneficial for the main task.
In \cite{lin2019adaptive}, gradient similarity is used to guide the adaptive selection of auxiliary tasks in reinforcement learning. In \cite{jiang2024forkmerge}, task weights are automatically adjusted by minimizing validation error. In AuxiLearn \cite{navon2020auxiliary}, nonlinear combinations of losses are learned dynamically to improve optimization. In AuxiNash \cite{shamsian2023auxiliary}, the AL problem is formulated as a generalized bargaining game  to resolve task interactions.

Unlike prior work, we propose a framework applicable to multiple auxiliary tasks, rather than methods tailored to a specific task. For example, Fan et al. \cite{fan2024bi} design a task-specific scheme using auxiliary image registration for one-shot segmentation, which is less flexible in terms of auxiliary task selection.
Sobotka et al. \cite{sobotka2024improving} require paired multimodal data as auxiliary labels, which falls outside our annotation-free setting.
Huang et al. \cite{huang2025fly} do not study auxiliary learning; instead, they introduce an auxiliary model for expert correction.

Registration-based methods have also been explored for label-efficient segmentation, including atlas-based unsupervised segmentation via contrastive registration that warps a single reference mask (CLMorph~\cite{liu2025clmorph}) and few-shot segmentation via self-supervised reconstruction (EFS-MedSeg~\cite{zhou2025efficient}); unlike these, ALBO neither relies on registration nor on unlabeled data, deriving its auxiliary labels for free from the existing main-task labels.

The optimization strategies above steer the auxiliary influence only through the loss or gradient of a shared encoder, by gating the auxiliary gradient with its similarity to the main gradient~\cite{du2018adapting,lin2019adaptive}, reweighting tasks by validation error~\cite{jiang2024forkmerge}, learning a nonlinear loss combination~\cite{navon2020auxiliary}, or bargaining over task weights~\cite{shamsian2023auxiliary}. They therefore neither model an explicit bidirectional exchange of features between the main and auxiliary branches, nor structurally guarantee that the main task is prioritized during optimization. ALBO addresses both limitations: the bidirectional communication module lets the main and auxiliary branches exchange features in both directions (Sec.~\ref{method:CCB}), and the bilevel formulation casts the main task as the upper level so the auxiliary task is optimized only insofar as it serves the main one (Sec.~\ref{method:PF}), using annotation-free auxiliary labels rather than task-specific extra data~\cite{fan2024bi,sobotka2024improving}.

\begin{figure*}[!t]
    \centering
    \includegraphics[width=1.0\linewidth]{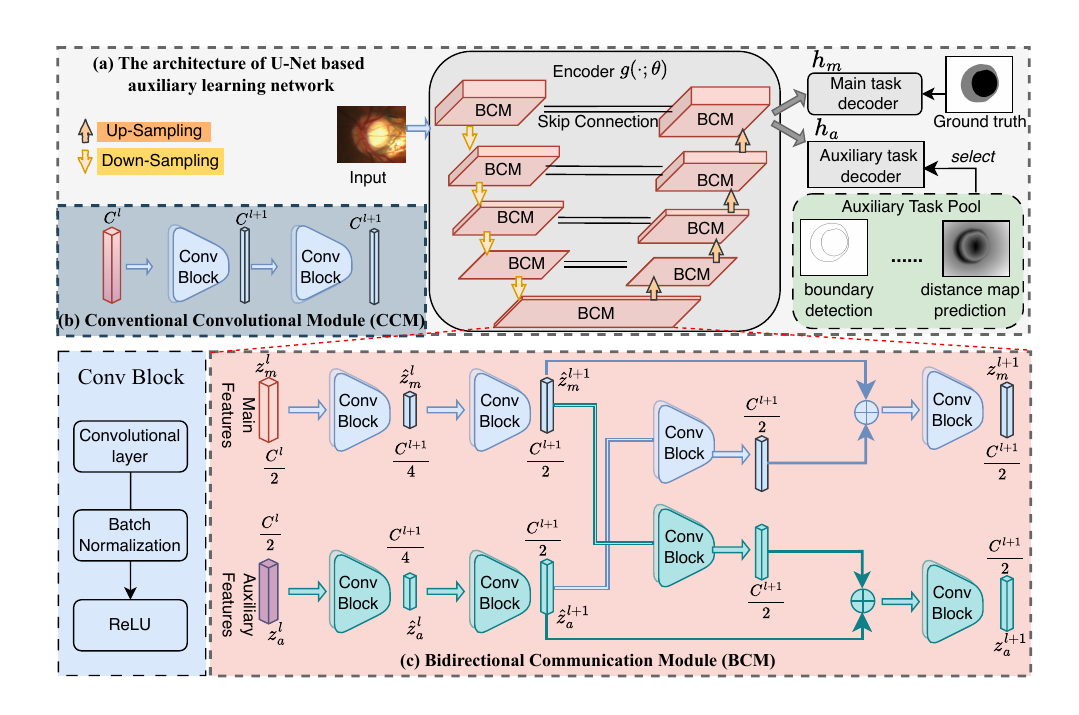}
    \caption{Overview of ALBO (auxiliary learning via bilevel optimization). (a) Network architecture: a U-Net-shaped shared encoder $g(\cdot;\theta)$ built from bidirectional communication modules (BCMs), followed by a main-task decoder $h_m$ for segmentation and an auxiliary-task decoder $h_a$. The auxiliary label (e.g., a boundary or distance map) is selected from an auxiliary task pool and is derived automatically from the ground-truth segmentation. All parameters are trained with the bilevel optimization (BLO) of Sec.~\ref{method:PF}. (b) Conventional convolutional module (CCM), i.e., two stacked Conv Blocks, used in place of the BCM in the ablation study. (c) BCM: the main (blue, top) and auxiliary (green, bottom) branches first learn task-specific features with two Conv Blocks each (step~1, Eq.~\eqref{step1:a}--\eqref{step1:b}), and then exchange features through cross-branch Conv Blocks fused by $\oplus$ (step~2, Eq.~\eqref{step2:e}--\eqref{step2:f}). Conv Block: convolution, batch normalization, and ReLU; $C^{l}$: number of channels at layer $l$.}
    \label{fig:framework}
\end{figure*}

\subsection{Multi-task learning (MTL)}\label{sec:related_mtl}
MTL, unlike AL, aims to improve the performance of all tasks simultaneously \cite{javaloy2021rotograd,ruder2017overview}, rather than prioritizing the main task.
Based on architectural design strategies, previous methods can be broadly categorized into soft parameter-sharing \cite{misra2016cross,ruder2019latent,wu2023multi} and hard parameter-sharing \cite{chen2018gradnorm,mgd_sener2018multi}.
Soft parameter-sharing methods design separate encoders for each task while enforcing constraints on their associated parameters. For example, MTAN \cite{liu2019end} employs an attention mechanism among task-specific networks to facilitate the sharing of a general feature pool.
Hard parameter-sharing methods, in contrast, primarily focus on finding a local update vector to reduce task conflict. 
Liu et al. introduce an approach to compute an update vector with equal projections on each task gradient \cite{liu2021towards}.
PCGrad \cite{pcgrad_yu2020gradient} projects each task’s gradient onto the normal plane of other task gradients, combining them to form the final update vector.
GradDrop \cite{graddrop_chen2020just} randomly drops task gradients based on the extent of their conflicts.
 CAGrad \cite{liu2021conflict} minimizes the average loss function and is theoretically proven to converge to the optimal point.
 Additionally, mixture of experts (MoE) architectures has also been widely adopted in MTL with an example of modeling the global task relationships by adding a generic convolution path to the original MoE structure \cite{yang2024multi}.

Beyond architectural design, a large body of MTL work targets the optimization of shared parameters, which can be grouped into loss-balancing schemes that reweight per-task losses, such as random loss weighting (RLW)~\cite{lin2021reasonable} and AMTL~\cite{senushkin2023independent}, and gradient-manipulation schemes that resolve conflicting task gradients, such as PCGrad~\cite{pcgrad_yu2020gradient}, GradDrop~\cite{graddrop_chen2020just}, CAGrad~\cite{liu2021conflict}, and Nash-MTL~\cite{navon2022multi}. However, these methods are designed for the symmetric MTL objective of improving all tasks jointly; they treat every task equally and do not privilege a designated main task, so a freely derived auxiliary signal can dilute rather than strengthen the primary segmentation. In contrast, ALBO adopts an asymmetric bilevel formulation that places the main segmentation task at the upper level and updates the auxiliary branch only to benefit it, which is better matched to the label-efficient auxiliary-learning setting studied here. We empirically compare against representative methods from both groups in Sec.~\ref{sec:comparison with al}.

\section{Methods}\label{method}

Because the auxiliary labels here are derived for free from the main-task data, our aim is not to solve the two tasks symmetrically but to make the auxiliary task \emph{serve} the main one; we therefore take AL as a bilevel problem in which the main task occupies the upper level.
Given an image $\mathbf{x}$, its corresponding main task label $\mathbf{y}_m$,  and an auxiliary label $\mathbf{y}_a$ automatically generated from $(\mathbf{x},\mathbf{y}_m)$, AL aims to learn a prediction model $f$ for the main task with the help of $\mathbf{y}_a$. Throughout the paper, we refer to $\mathbf{y}_a$ as the \emph{auxiliary label} and to the task of predicting it as the \emph{auxiliary task}.
Our model is composed of three functional modules: a \emph{shared encoder} $g(\cdot;\theta)$, a \emph{primary (main-task) head} $h_m(\cdot;\boldsymbol{\psi}_m)$, and an \emph{auxiliary head} $h_a(\cdot;\boldsymbol{\psi}_a)$, so that the main- and auxiliary-task predictors are $f=h_m\circ g$ and $f_a=h_a\circ g$, respectively. Because the encoder is built from the bidirectional communication module (BCM) in Sec.~\ref{method:CCB}, it carries two interacting branches, and its parameters split into a main branch $\boldsymbol{\theta}_m$ and an auxiliary branch $\boldsymbol{\theta}_a$, i.e., $\boldsymbol{\theta}=(\boldsymbol{\theta}_m,\boldsymbol{\theta}_a)$; the two heads carry parameters $\boldsymbol{\psi}_m$ and $\boldsymbol{\psi}_a$. Grouped by the role they play in optimization, the \textbf{main} (upper-level) variables are $\{\boldsymbol{\theta}_m,\boldsymbol{\theta}_a,\boldsymbol{\psi}_m\}$ and the \textbf{auxiliary} (lower-level) variables are $\{\boldsymbol{\theta}_a,\boldsymbol{\psi}_a\}$; the shared branch $\boldsymbol{\theta}_a$ is the variable through which the two levels interact.

As shown in Fig.~\ref{fig:framework}(a), in this work, our main task is to segment anatomical structures. We employ the U-Net \cite{ronneberger2015u} as the encoder.
To better exploit auxiliary supervision, we propose a novel \textbf{a}uxiliary \textbf{l}earning framework via \textbf{b}ilevel \textbf{o}ptimization, ALBO, which aligns with our motivation from both network architecture and optimization perspectives, as detailed in the following sections.

\subsection{Bidirectional Communication Module}
\label{method:CCB}

Generally, suppose that at the $l$-th layer of the encoder, we extract the main and auxiliary features from image $\mathbf{x}$, denoted by $\mathbf{z}^l_{m}\in \mathbb{R}^{\frac{C^{l}}{2}\times H\times W}$ and $\mathbf{z}^l_{a}\in \mathbb{R}^{\frac{C^{l}}{2}\times H\times W}$, respectively. $C^{l}$, $H$, and $W$ denote the number of channel, height, and width of the feature map, respectively.

To enable better auxiliary feature exploitation by the main task, we design the BCM to transfer knowledge from the auxiliary branch to the main one and vice versa, thereby exploiting the synergistic potential of reciprocally improving both tasks through information passing.
As shown in Fig.~\ref{fig:framework}(a), we replace the conventional convolutional module (CCM)  (Fig.~\ref{fig:framework}(b)) with the BCM (Fig.~\ref{fig:framework}(c)) as the core module of the encoder.
Specifically, given $\mathbf{z}^l_{m}$ and $\mathbf{z}^l_{a}$, we generate $\mathbf{z}^{l+1}_{m}$ and $\mathbf{z}^{l+1}_{a}$ at the $l$-th layer with the following two steps.
\begin{subequations}
    \begin{align}
    &\mathit{step}\ 1\ \textit{(Task-specific feature learning)}:   \notag\\
    &\hat{\mathbf{z}}^l_{m}= \mathrm{ConvB}(\mathbf{z}^l_{m}), \ \hat{\mathbf{z}}^l_{a}= \mathrm{ConvB}(\mathbf{z}^l_{a})\in \mathbb{R}^{\frac{C^{l+1}}{4}\times H\times W},\label{step1:a}\\
    &\hat{\mathbf{z}}^{l+1}_{m}= \mathrm{ConvB}(\hat{\mathbf{z}}^l_{m}), \hat{\mathbf{z}}^{l+1}_{a}= \mathrm{ConvB}(\hat{\mathbf{z}}^l_{a})\in \mathbb{R}^{\frac{C^{l+1}}{2}\times H\times W};\label{step1:b}
\end{align}
\end{subequations}
\begin{subequations}
    \begin{align}
    &\mathit{step}\ 2 \ \textit{(Feature communication)}: \notag\\
     &  \mathbf{z}^{l+1}_{m} = \mathrm{ConvB}(\hat{\mathbf{z}}^{l+1}_{m}\oplus \mathrm{ConvB}(\hat{\mathbf{z}}^{l+1}_{a}))\in \mathbb{R}^{\frac{C^{l+1}}{2}\times H\times W},\label{step2:e}\\
         & \mathbf{z}^{l+1}_{a} = \mathrm{ConvB}(\hat{\mathbf{z}}^{l+1}_{a}\oplus \mathrm{ConvB}(\hat{\mathbf{z}}^{l+1}_{m}))\in \mathbb{R}^{\frac{C^{l+1}}{2}\times H\times W},\label{step2:f}
\end{align}
\end{subequations}
where $\mathrm{ConvB}:=\mathrm{ReLU}\circ \mathrm{BN}\circ \mathrm{Conv}$ denotes convolutional block consisting of a convolutional layer ($\mathrm{Conv}$), a batch normalization ($\mathrm{BN}$), and a ReLU activation.

For simplicity, we denote the process from $\mathbf{z}^{l}_{m}$ to $\mathbf{z}^{l+1}_{m}$ as $g_m^{l+1}$, which is the composition of operations in Eqs.~\eqref{step1:a}, \eqref{step1:b}, and \eqref{step2:e} with learnable parameters $\boldsymbol{\theta}_m^{l+1}$, \textit{i.e.}, $\mathbf{z}^{l+1}_{m}=g_m^{l+1}(\mathbf{z}^{l}_{m}|\hat{\mathbf{z}}^{l+1}_{a};\boldsymbol{\theta}_m^{l+1})$.
Similarly, we have $\mathbf{z}^{l+1}_{a}=g_a^{l+1}(\mathbf{z}^{l}_{a}|\hat{\mathbf{z}}^{l+1}_{m};\boldsymbol{\theta}_a^{l+1})$ for the auxiliary branch with learnable parameters $\boldsymbol{\theta}_a^{l+1}$.


By setting $\mathbf{z}_m^0=\mathbf{z}_a^0=\mathbf{x}$, and using BCM in the encoder, the predictions of the main and auxiliary tasks become  
\begin{equation}
    \left\{ \begin{aligned}
        \hat{\mathbf{y}}_m &= h_{m}\circ g_m^{L}\circ \cdots g_m^{l}\cdots\circ g_m^{1}(\mathbf{x}) \\
        \hat{\mathbf{y}}_a &= h_{a}\circ g_a^{L}\circ \cdots g_a^{l}\cdots\circ g_a^{1}(\mathbf{x})
    \end{aligned}
    \right.,
\end{equation}
where $1\leq l\leq L$ is the encoding layer index.

\subsection{Formulation for AL using BCM}\label{method:PF}
When using BCM to build the encoder, the loss function for the main task is formulated as $\mathcal{L}_m(\boldsymbol{\theta}_m,\boldsymbol{\theta}_a,\boldsymbol{\psi}_m) = \mathbb{E}_{\mathbf{x}}[\ell_m(\hat{\mathbf{y}}_m,$ $\mathbf{y}_m)]$, and the loss function for the auxiliary task as $\mathcal{L}_a(\boldsymbol{\theta}_m,\boldsymbol{\theta}_a,\boldsymbol{\psi}_a) = \mathbb{E}_{\mathbf{x}}[\ell_a(\hat{\mathbf{y}}_a,\mathbf{y}_a)]$,
where $\ell_{m/a}$ measures the dissimilarity between the prediction and ground truth,
$\boldsymbol{\theta}_m=(\boldsymbol{\theta}_m^1,\cdots,\boldsymbol{\theta}_m^L)$ and $\boldsymbol{\theta}_a=(\boldsymbol{\theta}_a^1,\cdots,\boldsymbol{\theta}_a^L)$.
In this work, we adopt the cross-entropy loss \cite{sudre2017generalised} for $\ell_{m/a}$.

For AL, we consider the main task as our target, meaning we can combine the auxiliary and main branches as a whole for main task learning when minimizing $\mathcal{L}_m$, rather than treating the features from the auxiliary task as given conditions.
Hence, we formulate the AL problem as 
\begin{subequations}
    \begin{align}
        (\mathrm{main}):\quad\quad &\min_{\boldsymbol{\theta}_m,\boldsymbol{\theta}_a,\boldsymbol{\psi}_m} \mathcal{L}_m(\boldsymbol{\theta}_m,\boldsymbol{\theta}_a,\boldsymbol{\psi}_m), \label{eq:ucb:main}\\
        (\mathrm{auxiliary}):\quad & \min_{\boldsymbol{\theta}_a, \boldsymbol{\psi}_a}\mathcal{L}_a(\boldsymbol{\theta}_m, \boldsymbol{\theta}_a,\boldsymbol{\psi}_a).\label{eq:ucb:aux}
    \end{align}\label{eq:ucb}
\end{subequations}

To solve the above bilevel optimization problem, we employ an approximation scheme by alternately updating the parameters.
Specifically, 
we have
\begin{subequations}
    \begin{align}
        \boldsymbol{\theta}_a^{'} &\leftarrow \boldsymbol{\theta}_a - \alpha\triangledown_{\boldsymbol{\theta}_a}\mathcal{L}_a(\boldsymbol{\theta}_m,\boldsymbol{\theta}_a,\boldsymbol{\psi}_a),\\
        [\boldsymbol{\theta}_{m/a},\boldsymbol{\psi}_m] &\leftarrow [\boldsymbol{\theta}_{m/a},\boldsymbol{\psi}_m] - \beta \triangledown_{[\boldsymbol{\theta}_{m/a},\boldsymbol{\psi}_m]}\mathcal{L}_m(\boldsymbol{\theta}_m,\boldsymbol{\theta}_a^{'},\boldsymbol{\psi}_m),\label{eq:bcb:grad_m}\\
        [\boldsymbol{\theta}_a,\boldsymbol{\psi}_a] &\leftarrow [\boldsymbol{\theta}_a,\boldsymbol{\psi}_a] - \alpha\triangledown_{[\boldsymbol{\theta}_a,\boldsymbol{\psi}_a]}\mathcal{L}_a(\boldsymbol{\theta}_m,\boldsymbol{\theta}_a,\boldsymbol{\psi}_a),
    \end{align}\label{eq:bcb:grad}
\end{subequations}
where $\alpha$ and $\beta$ are learning rates.

\emph{Optimization intuition.} Rather than minimizing $\mathcal{L}_m$ at the current auxiliary branch, the main update Eq.~\eqref{eq:bcb:grad_m} evaluates it at the \emph{look-ahead} branch $\boldsymbol{\theta}_a^{'}=\boldsymbol{\theta}_a-\alpha\triangledown_{\boldsymbol{\theta}_a}\mathcal{L}_a$, i.e., one gradient step ahead on the auxiliary task. 
Differentiating through this step introduces a second-order coupling term, $-\alpha\,\triangledown^2_{\boldsymbol{\theta}_m,\boldsymbol{\theta}_a}\mathcal{L}_a\cdot\triangledown_{\boldsymbol{\theta}_a^{'}}\mathcal{L}_m$. This term pairs the sensitivity of the main loss to the shared auxiliary parameters, $\triangledown_{\boldsymbol{\theta}_a^{'}}\mathcal{L}_m$, with the curvature of the auxiliary objective. It thus measures, to first order, how the one-step auxiliary update perturbs the shared representation. This perturbation, in turn, alters the main objective.
Intuitively, it reinforces auxiliary updates whose direction agrees with the main-task descent and discounts those that would increase $\mathcal{L}_m$, providing a built-in safeguard against \emph{negative transfer}. This is the mechanism by which ALBO utilizes the auxiliary task to benefit, rather than compete with, the main task.

The outer update Eq.~\eqref{eq:bcb:grad_m} requires the second-order gradient $\triangledown_{\boldsymbol{\theta}_{m/a}}\mathcal{L}_m(\boldsymbol{\theta}_m,\boldsymbol{\theta}_a^{'},\boldsymbol{\psi}_m)$, which is memory- and time-consuming. To avoid explicitly forming the Hessian, we adopt a finite-difference approximation~\cite{liu2022auto} that needs only two additional forward--backward passes, giving the final update gradient
\begin{equation}
    \begin{aligned}
        &\triangledown_{\boldsymbol{\theta}_m/\boldsymbol{\theta}_a^{'}}\mathcal{L}_m(\boldsymbol{\theta}_m,\boldsymbol{\theta}_a^{'},\boldsymbol{\psi}_m)\approx \triangledown_{\boldsymbol{\theta}_m/\boldsymbol{\theta}_a^{'}}\mathcal{L}_m(\boldsymbol{\theta}_m,\boldsymbol{\theta}_a^{'},\boldsymbol{\psi}_m)\\
        &\quad-\alpha \frac{\triangledown_{\boldsymbol{\theta}_m/\boldsymbol{\theta}_a^{+}}\mathcal{L}_a(\boldsymbol{\theta}_m,\boldsymbol{\theta}_a^{+},\boldsymbol{\psi}_a)-\triangledown_{\boldsymbol{\theta}_m/\boldsymbol{\theta}_a^{-}}\mathcal{L}_a(\boldsymbol{\theta}_m,\boldsymbol{\theta}_a^{-},\boldsymbol{\psi}_a)}{2\epsilon},
    \end{aligned}\label{eq:fid}
\end{equation}
where $\boldsymbol{\theta}_a^{\pm}=\boldsymbol{\theta}_a \pm \epsilon\cdot\triangledown_{\boldsymbol{\theta}_a^{'}}\mathcal{L}_m(\boldsymbol{\theta}_m,\boldsymbol{\theta}_a^{'},\boldsymbol{\psi}_m)$ and $\epsilon$ is a small positive scalar. The complete step-by-step derivation of Eq.~\eqref{eq:fid} is given in Appendix~\ref{app:fd}.

In terms of complexity, this finite-difference scheme computes the required Hessian--vector product using only first-order gradients (one look-ahead step and two extra forward--backward passes per outer iteration), so each iteration retains the $O(P)$ time and memory of single-level training in the number of shared parameters $P$, avoiding both the $O(P^2)$ explicit Hessian and the heavier second-order automatic differentiation; the bilevel update thus adds only a constant-factor overhead, confined to training.

\begin{table*}[t!]

\caption{Ablation study on the REFUGE 2018 dataset using U-Net as the backbone with 128 (U-Net-S) bottom channels. In the `BCM' column, \xmark~means using the CCM module instead. In `BLO' column, \xmark~(\textit{Method}) means using \textit{Method} to train the model instead of BLO. OC: optic cup; OD: optic disc. Best and second-best values in each column are highlighted in \textcolor{best}{red} and \textcolor{second}{blue}, respectively.
}\label{tab:ablation}
\centering
\setlength{\tabcolsep}{12pt}
\scalebox{1.0}{
\begin{tabular}{clcccccccc}
\toprule
\multirow{3}{*}{BCM}&\multirow{3}{*}{BLO}& \multicolumn{4}{c}{Auxiliary task: Boundary detection}& \multicolumn{4}{c}{Auxiliary task: Distance-map prediction}\\
\cmidrule(lr){3-6}\cmidrule(lr){7-10}
			& &\multicolumn{2}{c}{Dice (\%) $\uparrow$}  & \multicolumn{2}{c}{ASSD (pixels) $\downarrow$} &\multicolumn{2}{c}{Dice (\%) $\uparrow$}  & \multicolumn{2}{c}{ASSD (pixels) $\downarrow$} \\
			\cmidrule(lr){3-4}\cmidrule(lr){5-6}\cmidrule(lr){7-8}\cmidrule(lr){9-10}
		& &	OC& OD   & OC & OD  &	OC& OD   & OC & OD \\
\midrule
 \xmark & \xmark~(\textit{Independent}) &74.77 & 91.18&21.94 &14.51 & 74.77 & 91.18&21.94 &14.51 \\

  \xmark & \xmark~(\textit{Joint}) &74.65 & 90.93&22.08 &13.79 & 74.46 & 90.55&20.72 &13.84 \\

\midrule
 \cmark  & \xmark~(\textit{Independent})    &  81.54 &  \textcolor{second}{93.17}  & 10.28   &  \textcolor{second}{\ 7.74}  & \textcolor{second}{81.54} &  \textcolor{second}{93.17}  & \textcolor{second}{10.28}   &  \textcolor{second}{\ 7.74} \\

  \cmark  &\xmark~(\textit{Joint}) &  \textcolor{best}{82.91} &  93.09  & \textcolor{second}{\ 9.62}   &  \ 7.86  &81.41 &93.00 &10.90 &\ 7.87 \\

\midrule
 
\xmark & \cmark & 77.73&88.97 &14.71 &17.58 &78.29 &91.44 &14.06 & 11.90 \\
\midrule
 \multicolumn{2}{c}{ALBO (Ours) }   & \textcolor{second}{82.30}  &  \textcolor{best}{94.32} & \textcolor{best}{\ 9.36}   &  \textcolor{best}{\ 6.38} &\textcolor{best}{83.99} & \textcolor{best}{93.52}&\textcolor{best}{\ 8.90} &\textcolor{best}{\ 7.34}  \\

\bottomrule
\end{tabular}
}

\end{table*}

\begin{figure}[t!]
\centering
\includegraphics[width=\linewidth]{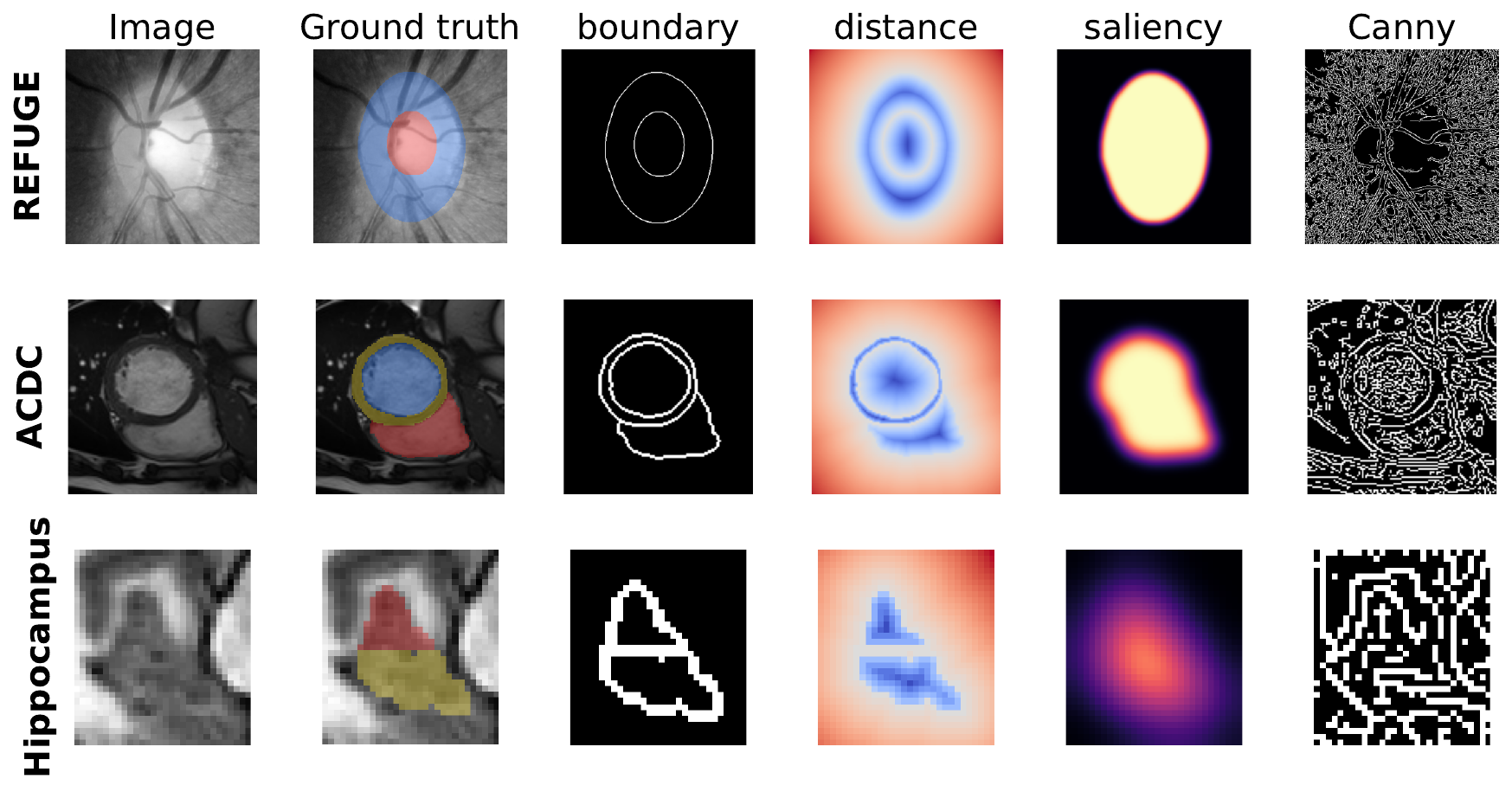}
\caption{The four auxiliary-task labels used in our study, shown for one representative sample per dataset (REFUGE, ACDC, Hippocampus). From left to right: input image, ground-truth segmentation, and the boundary, signed-distance, saliency, and Canny-edge auxiliary labels. The boundary and distance labels are derived from the segmentation mask, the saliency label is its Gaussian-smoothed foreground, and the Canny label is derived from the image intensity.}
\label{fig:aux_labels}
\end{figure}

\section{Experiments}

\subsection{Datasets and pre-processing}

We evaluated the proposed method with three segmentation tasks: (1) Optic cup/disc (OC/OD) segmentation using the REFUGE 2018 dataset \cite{ORLANDO2020101570}. 
REFUGE consists of 400 color fundus photography images for training, 400 for validation, and 400 for testing.
All images were cropped to a size of $512 \times 512$.
For REFUGE, each model was trained \textit{three times} with different random initializations to ensure reliability.
(2) Cardiac segmentation using the ACDC dataset \cite{bernard2018deep}, which contains 200 short-axis cardiac MR-cine images with the labels of the right ventricle (RV), myocardium (Myo), and left ventricle (LV).
All slices were cropped to a size of $128 \times 128$.
For ACDC, each model was evaluated with \textit{five-fold cross-validation}.
For each fold, we used 40 cases for validation, and the remaining 160 cases for the training data selection.
(3) The Hippocampus dataset from Medical Segmentation Decathlon Challenge \cite{simpson2019large} consists of 260 3D scans with annotations of the hippocampus head (HH) and body (HB) available.
All volumes were resized to $48\times 64 \times 48$ with zero-padding.
For the Hippocampus, we also employed \textit{five-fold cross-validation}.
For each fold, we used 52 scans for validation, and the remaining 208 scans for the training set selection.
For evaluation, the average Dice and the average symmetric surface distance (ASSD) were reported.

\subsection{Implementation details}

We implemented nine representative methods for comparison:
(1) Independent: this is the conventional independent learning for the main segmentation tasks without involving any auxiliary tasks;
(2) Four AL methods, including Joint, GCS \cite{du2018adapting}, AuxiLearn \cite{navon2020auxiliary}, and AuxiNash \cite{shamsian2023auxiliary}.  
(3) Four MTL methods, including CAGrad \cite{liu2021conflict}, Nash \cite{navon2022multi}, RLW \cite{lin2021reasonable}, and AMTL \cite{senushkin2023independent}.
Concretely, \emph{Independent} trains only the main task, minimizing $\mathcal{L}_m(\boldsymbol{\theta}_m,\boldsymbol{\theta}_a,\boldsymbol{\psi}_m)$ over $\{\boldsymbol{\theta}_m,\boldsymbol{\theta}_a,\boldsymbol{\psi}_m\}$ with no auxiliary loss, whereas \emph{Joint} minimizes the equally-weighted sum $\mathcal{L}_m+\mathcal{L}_a$ over all parameters $\{\boldsymbol{\theta}_m,\boldsymbol{\theta}_a,\boldsymbol{\psi}_m,\boldsymbol{\psi}_a\}$ by conventional single-level gradient descent, i.e., without the bilevel look-ahead of ALBO.
For comparison, all models were evaluated using the same U-Net-shaped network (as shown in Fig.~\ref{fig:framework}(a)), and run on a single 40GB NVIDIA A100 GPU. 
We employed 3D models for the Hippocampus and 2D models for the others.
For simplicity, we denote the U-Net with 128 and 1024 bottom channels as U-Net-S and U-Net-L, respectively. For all datasets, we trained ALBO using the SGD optimizer with the initial learning rate of $\alpha=\beta=0.1$ and decay of 0.9 every 200 iterations.
All models were trained for a maximum of $15,000$ iterations with a batch size of 20.

\begin{table*}[t]
\caption{Effect of auxiliary-task selection and combination, with 10 labeled images. All variants use the identical ALBO bilevel optimizer and backbone; only the auxiliary set changes. 
Best and second-best per column are in \textcolor{best}{red} and \textcolor{second}{blue}.}
\label{tab:aux_ablation}
\centering
\setlength{\tabcolsep}{5pt}
\resizebox{0.8\textwidth}{!}{
\begin{tabular}{lcccccccccc}
\toprule
\multirow{3}{*}{Auxiliary task(s)} & \multicolumn{4}{c}{\textbf{REFUGE (3 seeds)}} & \multicolumn{6}{c}{\textbf{ACDC (5 folds)}} \\
\cmidrule(lr){2-5}\cmidrule(lr){6-11}
 & \multicolumn{2}{c}{Dice (\%)$\uparrow$} & \multicolumn{2}{c}{ASSD (pixels)$\downarrow$} & \multicolumn{3}{c}{Dice (\%)$\uparrow$} & \multicolumn{3}{c}{ASSD (mm)$\downarrow$} \\
\cmidrule(lr){2-3}\cmidrule(lr){4-5}\cmidrule(lr){6-8}\cmidrule(lr){9-11}
 & OC & OD & OC & OD & RV & Myo & LV & RV & Myo & LV \\
\midrule
\multicolumn{11}{l}{\textit{Single auxiliary task}} \\
Boundary          & \textcolor{best}{73.58} & \textcolor{second}{82.42} & 18.76 & 28.32 & \textcolor{best}{70.74} & \textcolor{best}{80.21} & \textcolor{best}{87.48} & \textcolor{best}{4.29}  & \textcolor{best}{1.49} & \textcolor{best}{2.19} \\
Distance   & 65.36 & 78.28 & 24.31 & 34.62 & 68.46 & 78.68 & 86.23 & 5.03 & 1.79 & 2.55 \\
Saliency          & 67.71 & 81.73 & 18.71 & \textcolor{second}{27.03} & 65.96 & 76.87 & 85.32 & 5.53 & 1.83 & 2.96 \\
Canny             & 66.24 & 78.57 & 20.50 & 33.72 & 69.66 & \textcolor{second}{79.98} & \textcolor{second}{87.22} & 4.60 & \textcolor{second}{1.40} & 2.27 \\
\midrule
\multicolumn{11}{l}{\textit{Two auxiliary tasks}} \\
Boundary+Canny    & 72.18 & 80.89 & 18.74 & 29.45 & \textcolor{second}{70.36} & 78.99 & 87.14 & 5.13 & 1.54 & 2.20 \\
Boundary+Distance      & 66.82 & 75.97 & 24.67 & 38.75 & 68.08 & 76.34 & 84.43 & 5.27 & 1.88 & 2.68 \\
\midrule
\multicolumn{11}{l}{\textit{Three auxiliary tasks}} \\
Boundary+Canny+Distance        & 68.42 & 79.60 & \textcolor{second}{18.52} & 31.29 & 68.10 & 79.47 & 86.85 & \textcolor{second}{4.41} & 1.56 & \textcolor{second}{2.18} \\
Boundary+Distance+Saliency     & 68.10 & 81.34 & 19.11 & 27.03 & 66.53 & 75.57 & 83.88 & 4.98 & 1.97 & 3.08 \\
\midrule
\multicolumn{11}{l}{\textit{Four auxiliary tasks}} \\
Boundary+Distance+Saliency+Canny & \textcolor{second}{72.81} & \textcolor{best}{83.49} & \textcolor{best}{16.02} & \textcolor{best}{23.67} & 66.85 & 76.17 & 84.59 & 4.76 & 2.00 & 2.75 \\
\bottomrule
\end{tabular}
}
\end{table*}

\subsection{Ablation study}
We conducted an ablation study to evaluate the effectiveness of the two key components of ALBO, \textit{i.e.,} the BCM module and the bilevel optimization (BLO) algorithm, using the whole REFUGE dataset.
Since the effectiveness of different automatically generated auxiliary tasks is unclear, we evaluated two widely used tasks, \textit{i.e.,} boundary detection and distance-map prediction \cite{luo2021semi}, for comparison.

As presented in Table \ref{tab:ablation}, the proposed method achieved the best results across most metrics when both BCM and BLO are enabled, and remained robust across different auxiliary tasks.
Removing BLO resulted in a noticeable performance degradation for both auxiliary tasks. 
For example, with distance-map prediction as the auxiliary task, the OC Dice decreased from $83.99\%$ to $81.41\%$, while with boundary detection the OD Dice decreased from $94.32\%$ to $93.09\%$.
Without BCM, the parameters $\boldsymbol{\theta}_m$ and $\boldsymbol{\theta}_a$ in Eq. \eqref{eq:ucb} merged into a shared encoder, \textit{i.e.,} $\boldsymbol{\theta}$, and the performance of the penultimate row degraded significantly (paired t-test: $p<0.01$).
In this configuration the bilevel update acts on a fully shared encoder without a dedicated auxiliary branch, so the boundary auxiliary task is imposed directly on the shared main-task features. This can over-constrain the higher-contrast optic disc representation, reducing its Dice, while the smaller optic cup still benefits from the auxiliary signal. These results suggest that BLO is most effective when combined with BCM, since branch separation and bidirectional communication help route the auxiliary signal without causing negative transfer.

Comparing the first four rows of Table~\ref{tab:ablation} further shows that BCM alone improves the performance of Independent and Joint on REFUGE.
However, it does not mean that BCM alone can boost the performance in all cases.
For example, experiments show that on the ACDC dataset, all  nine compared methods obtained inferior results when using the BCM module instead of the CCM module.
Remarkably, the utility of BCM can be exploited robustly by BLO. 
More detailed discussion about the effect of BCM and CCM can be found in Appendix \ref{app:bcm}.

\subsection{Effect of auxiliary-task selection and combination}
\label{sec:aux_ablation}

We study how the choice and number of auxiliary tasks affect segmentation performance by evaluating four auxiliary tasks, namely boundary, distance, saliency, and Canny edge (Fig.~\ref{fig:aux_labels}), both individually and in combination.

Table~\ref{tab:aux_ablation} reports results with 10 labeled samples from REFUGE and ACDC. 
First, the boundary task is the strongest and most robust single auxiliary task: it attained the best score on all six ACDC metrics and the best OC Dice on REFUGE, and no other single task was competitive across both datasets. 
The remaining single tasks were weaker and less consistent.
The Canny edge was competitive on ACDC but markedly worse on REFUGE, while the distance and saliency tasks were the weakest overall. 
Second, combining auxiliary tasks was not systematically beneficial. On REFUGE, using all four tasks gave the best OD Dice and the lowest ASSD, marginally ahead of the boundary task; on ACDC, however, every combination was inferior to the single boundary task, and the four-task set fell well below it.

These results indicate that the bilevel feature communication mechanism extracts most of the informative structure from a single, well-matched auxiliary task. 
Adding more tasks provides, at best, a marginal and dataset-dependent gain, and may harm accuracy.
Thus, increasing the number of auxiliary tasks is not intrinsically beneficial. This justifies our choice of a single, mask-derived boundary auxiliary task: it is the most robust option across both modalities, and avoids the task-balancing burden and negative-transfer risk of multi-auxiliary training.

\begin{table*}[th!]

\caption{Performance comparison on the REFUGE dataset with 400 and 10 labeled images for training. 
The number of parameters (\# Params) for each model is around ~0.5M.
Best and second-best values in each column are shown in \textcolor{best}{red} and \textcolor{second}{blue}, respectively.
Each ASSD (pixels) value is followed by the 95\% Hausdorff Distance (HD95, pixels) in parentheses. $^{*}$ and $^{**}$ indicate a statistically significant difference from ALBO with $p<0.05$ and $p<0.01$, respectively (paired Wilcoxon signed-rank test on the per-image scores averaged over the three seeds).
}\label{tab:refuge_effi}
\centering
\setlength{\tabcolsep}{6pt}
\resizebox{\textwidth}{!}{
\begin{tabular}{lcccccccc}
\toprule
\multirow{3}{*}{Method} &
\multicolumn{4}{c}{\textbf{400 labeled images}} &
\multicolumn{4}{c}{\textbf{10 labeled images}} \\
\cmidrule(lr){2-5}\cmidrule(lr){6-9}
 & \multicolumn{2}{c}{\small Dice (\%)$\uparrow$} & \multicolumn{2}{c}{\small ASSD (HD95) $\downarrow$} & \multicolumn{2}{c}{\small Dice (\%)$\uparrow$} & \multicolumn{2}{c}{\small ASSD (HD95) $\downarrow$} \\
 \cmidrule(lr){2-3}\cmidrule(lr){4-5}\cmidrule(lr){6-7}\cmidrule(lr){8-9}
 & OC& OD   & OC & OD &	OC& OD   & OC & OD   \\
\midrule

Independent+BCM & 81.54$^{**}$ & 93.17$^{**}$ & 10.28$^{**}$ (23.24$^{**}$) & \ 7.74$^{**}$ (18.82$^{**}$) & \textcolor{second}{67.54}$^{**}$ & \textcolor{second}{80.91}$^{**}$ & 22.83$^{**}$ (56.33$^{**}$) & \textcolor{second}{30.50}$^{**}$ (\textcolor{second}{89.08}) \\
Joint+BCM & \textcolor{second}{82.91} & 93.09$^{**}$ & \ 9.62 (21.59$^{**}$) & \ 7.86$^{**}$ (18.18$^{**}$) & 65.01$^{**}$ & 77.33$^{**}$ & 23.93$^{**}$ (56.42$^{**}$) & 36.07$^{**}$ (92.71$^{**}$) \\
GCS+BCM \cite{du2018adapting} & \textcolor{best}{83.39}$^{**}$ & \textcolor{second}{93.78}$^{**}$ & \textcolor{best}{\ 9.30}$^{*}$ (\textcolor{second}{20.51}) & \textcolor{second}{\ 6.98}$^{**}$ (\textcolor{second}{16.07}$^{**}$) & 64.75$^{**}$ & 77.30$^{**}$ & 26.66$^{**}$ (60.92$^{**}$) & 36.74$^{**}$ (97.79$^{**}$) \\
AuxiLearn+BCM \cite{navon2020auxiliary} & 82.75 & 93.32$^{**}$ & \ 9.93 (22.72$^{**}$) & \ 7.53$^{**}$ (18.64$^{**}$) & 63.53$^{**}$ & 78.11$^{**}$ & 27.49$^{**}$ (64.67$^{**}$) & 36.50$^{**}$ (91.98$^{**}$) \\
AuxiNash +BCM  \cite{shamsian2023auxiliary} & 81.37$^{**}$ & 93.07$^{**}$ & 10.36$^{**}$ (23.38$^{**}$) & \ 7.81$^{**}$ (19.75$^{**}$) & 59.74$^{**}$ & 73.52$^{**}$ & 38.84$^{**}$ (87.78$^{**}$) & 44.29$^{**}$ (111.63$^{**}$) \\
RLW+BCM \cite{lin2021reasonable} & 81.05$^{**}$ & 93.42$^{**}$ & 10.31$^{**}$ (22.61$^{**}$) & \ 7.44$^{**}$ (18.77$^{**}$) & 67.48$^{**}$ & 73.86$^{**}$ & \textcolor{second}{20.17}$^{**}$ (\textcolor{best}{47.79}) & 43.89$^{**}$ (101.17$^{**}$) \\
Nash+BCM \cite{navon2022multi} & 80.97$^{**}$ & 93.11$^{**}$ & 10.76$^{**}$ (23.93$^{**}$) & \ 7.84$^{**}$ (19.40$^{**}$) & 63.64$^{**}$ & 74.24$^{**}$ & 24.93$^{**}$ (63.93$^{**}$) & 42.37$^{**}$ (110.77$^{**}$) \\
AMTL+BCM \cite{senushkin2023independent} & 81.16$^{**}$ & 93.66$^{**}$ & 10.49$^{**}$ (23.68$^{**}$) & \ 7.07$^{**}$ (17.74$^{**}$) & 63.61$^{**}$ & 77.94$^{**}$ & 24.13$^{**}$ (59.05$^{**}$) & 35.77$^{**}$ (101.58$^{**}$) \\
CAGrad+BCM  \cite{liu2021conflict} & 82.55 & 93.62$^{**}$ & \ 9.60 (21.15$^{**}$) & \ 7.15$^{**}$ (16.98$^{**}$) & 67.20$^{**}$ & 79.77$^{**}$ & 22.23$^{**}$ (54.56$^{**}$) & 31.25$^{**}$ (92.15$^{**}$) \\
\midrule
ALBO (Ours) & 82.30 & \textcolor{best}{94.32} & \textcolor{second}{\ 9.36} (\textcolor{best}{19.81}) & \textcolor{best}{\ 6.38} (\textcolor{best}{15.44}) & \textcolor{best}{73.58} & \textcolor{best}{82.42} & \textcolor{best}{18.76} (\textcolor{second}{48.76}) & \textcolor{best}{28.32} (\textcolor{best}{87.66}) \\

\bottomrule
\end{tabular}
}

\end{table*}

\begin{figure*}[t]
  \centering
  \begin{subfigure}{0.54\linewidth}
    \centering
    \includegraphics[width=\linewidth]{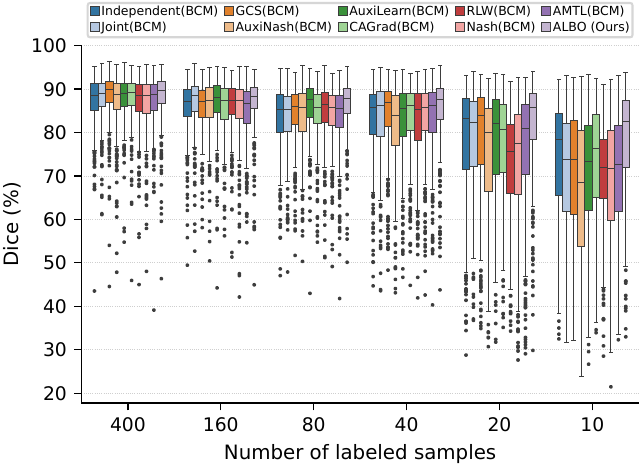}
    \caption{Comparison on REFUGE.}
    \label{fig:acdc_data_eff-a}
  \end{subfigure}
  \hfill
  \begin{subfigure}{0.44\linewidth}
    \centering
    \includegraphics[width=\linewidth]{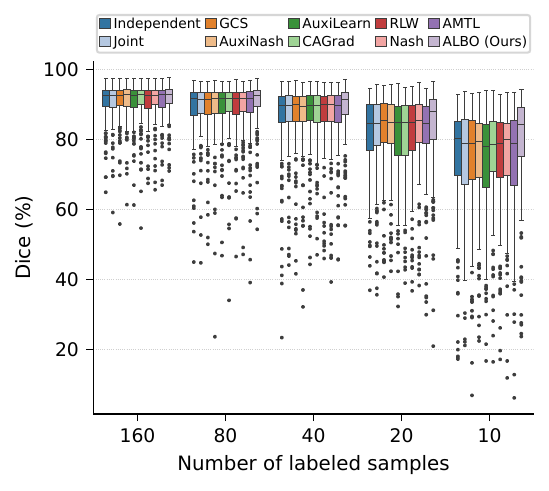}
    \caption{Comparison on ACDC.}
    \label{fig:acdc_data_eff-b}
  \end{subfigure}
  \caption{Performance change of all compared methods on the REFUGE and ACDC with varying numbers of training samples.}
  \label{fig:acdc_data_eff}
\end{figure*}

\begin{figure*}[t!]
    \centering
    \includegraphics[width=1.0\linewidth]{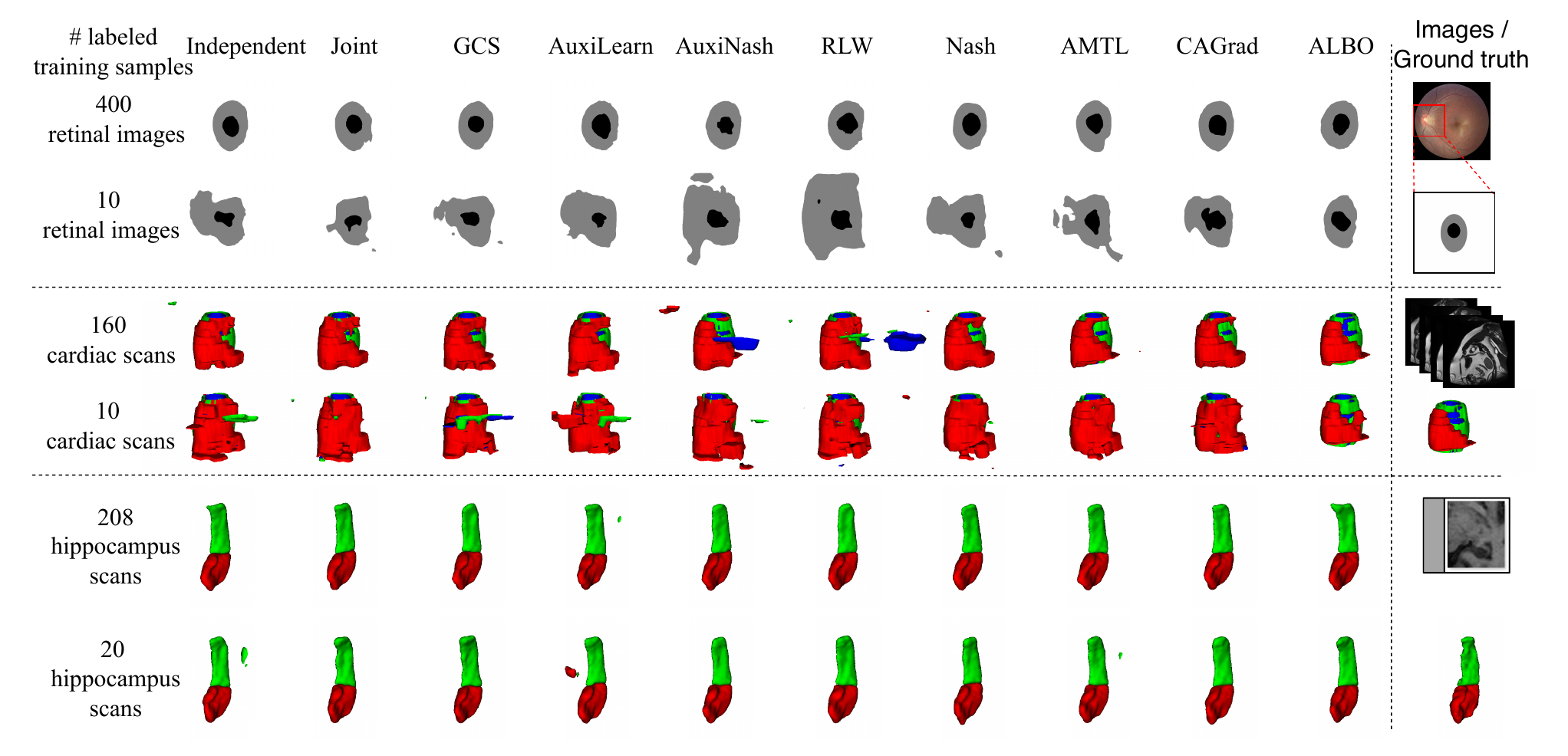}
    \caption{Visualization of segmentation predictions from compared methods trained with different numbers of labeled samples. 
    Here, each sample for visualization is the median case of the Independent method (trained with full data) in terms of Dice.
    }
    \label{fig:Visual_comparison}
\end{figure*}

\subsection{Label efficiency study}\label{sec:comparison with al}
We evaluated the label efficiency of ALBO by training the model with different numbers of samples and using boundary detection as the auxiliary task.

\subsubsection{Comparison on REFUGE.}
For OC and OD segmentation, we randomly selected 10, 20, 40, 80, and 160 retinal images for training.
Since all compared methods achieved significantly better performance ($p<0.01$) with the BCM than with the CCM when training with the whole 400 labeled images, we used BCM-embedded U-Net-S for all methods to provide a fairer comparison of label efficiency.

Table \ref{tab:refuge_effi} presents the average performance on OC and OD of all methods, trained with either 400 or 10 labeled images.
With sufficient training data, the performance of all methods was close.
When the labeled data were reduced to 10 samples, ALBO demonstrated the best robustness, whereas all other AL and MTL methods performed worse than Independent.
We additionally report the 95\% Hausdorff Distance (HD95) and, using a paired Wilcoxon signed-rank test, mark in Table~\ref{tab:refuge_effi} the metrics that differ significantly from ALBO. With 400 labeled images, although the Dice scores of all methods were similar, ALBO attained the best HD95 on both OC and OD, indicating more accurate boundary delineation. With only 10 labeled images, ALBO yielded the best Dice and ASSD on both structures, and its Dice and ASSD improvements over the competing methods were statistically significant ($p<0.01$).

Fig.~\ref{fig:acdc_data_eff} (a) illustrates the trend of Dice scores averaged on OC and OD obtained by all methods as the number of labeled training samples decreased.
Overall, ALBO is more robust to changes in training data size.
When the number of labeled training samples is $N\leq 80$, ALBO significantly outperforms the competing methods in average Dice ($p<0.05$).
A similar trend can be observed for ASSD.
The top two rows of Fig.~\ref{fig:Visual_comparison} illustrate  predictions from each method on a representative test sample, trained with 400 and 10 retinal images, respectively.
These results show that ALBO achieves the best performance when the labeled training data are limited.

\begin{table*}[tb]
\caption{Performance comparison on the ACDC dataset using 160 and 10 labeled scans for training.
Best and second-best values in each column are shown in \textcolor{best}{red} and \textcolor{second}{blue}, respectively.
Each ASSD (mm) value is followed by the 95\% Hausdorff Distance (HD95, mm) in parentheses. $^{*}$ and $^{**}$ indicate a statistically significant difference from ALBO with $p<0.05$ and $p<0.01$, respectively (paired Wilcoxon signed-rank test over all samples).
}\label{tab:acdc_effi}
\centering

\setlength{\tabcolsep}{3pt}
\scalebox{0.85}{
\begin{tabular}{lcccccccccccc}
\toprule
\multirow{3}{*}{Method}&\multicolumn{6}{c}{\textbf{160 labeled scans}}&\multicolumn{6}{c}{\textbf{10 labeled scans}}\\
\cmidrule(lr){2-7}\cmidrule(lr){8-13}
 &\multicolumn{3}{c}{Dice (\%) $\uparrow$}  & \multicolumn{3}{c}{ASSD (HD95) $\downarrow$} &\multicolumn{3}{c}{Dice (\%) $\uparrow$}  & \multicolumn{3}{c}{ASSD (HD95)  $\downarrow$} \\
			\cmidrule(lr){2-4}\cmidrule(lr){5-7}\cmidrule(lr){8-10}\cmidrule(lr){11-13}
			&	RV& Myo & LV   & RV& Myo & LV &	RV& Myo & LV   & RV& Myo & LV  \\
 \hline
Independent & 89.16$^{**}$ & 89.47$^{**}$ & 94.01$^{*}$ & 1.03$^{*}$ (\textcolor{best}{5.32}) & 0.44$^{*}$ (2.47) & 0.63$^{*}$ (3.46) & 67.92$^{**}$ & 74.97$^{**}$ & 83.22$^{**}$ & \textcolor{second}{5.75}$^{**}$ (\textcolor{second}{22.87}$^{**}$) & \textcolor{second}{2.21}$^{**}$ (\textcolor{second}{10.58}$^{**}$) & 3.01$^{**}$ (12.95$^{**}$) \\
Joint & 88.93$^{**}$ & 89.56$^{**}$ & 94.19 & 1.11$^{**}$ (6.01$^{*}$) & 0.45$^{**}$ (2.42) & 0.63 (3.44) & 66.22$^{**}$ & 75.67$^{**}$ & 81.98$^{**}$ & 6.25$^{**}$ (25.79$^{**}$) & 2.55$^{**}$ (13.71$^{**}$) & 3.42$^{**}$ (14.69$^{**}$) \\
GCS  \cite{du2018adapting} & 89.20$^{**}$ & 89.49$^{**}$ & 94.14$^{**}$ & 1.04$^{**}$ (5.51) & 0.46$^{**}$ (2.43$^{*}$) & 0.61$^{**}$ (3.39) & 66.70$^{**}$ & 75.14$^{**}$ & 82.08$^{**}$ & 6.35$^{**}$ (27.03$^{**}$) & 2.71$^{**}$ (13.14$^{**}$) & 3.60$^{**}$ (15.52$^{**}$) \\
AuxiLearn \cite{navon2020auxiliary} & 89.09$^{**}$ & 89.47$^{**}$ & 94.15 & 1.03 (5.46) & 0.45$^{*}$ (2.37) & 0.56 (\textcolor{second}{2.97}) & 65.53$^{**}$ & 74.00$^{**}$ & 80.78$^{**}$ & 6.82$^{**}$ (27.46$^{**}$) & 2.55$^{**}$ (12.71$^{**}$) & 3.38$^{**}$ (13.72$^{**}$) \\
AuxiNash  \cite{shamsian2023auxiliary} & 89.06$^{**}$ & \textcolor{second}{89.85} & 94.31 & 1.14$^{*}$ (6.21$^{*}$) & \textcolor{second}{0.42} (2.29) & \textcolor{second}{0.54} (3.07) & 66.86$^{**}$ & 75.62$^{**}$ & 82.37$^{**}$ & 6.96$^{**}$ (29.23$^{**}$) & 2.41$^{**}$ (11.95$^{**}$) & 3.24$^{**}$ (14.24$^{**}$) \\
RLW  \cite{lin2021reasonable} & 89.18$^{**}$ & 89.43$^{**}$ & 94.22 & 1.04$^{**}$ (5.65) & 0.45$^{*}$ (2.56) & 0.59 (3.51) & 67.33$^{**}$ & 74.58$^{**}$ & 82.38$^{**}$ & 6.43$^{**}$ (27.37$^{**}$) & 2.55$^{**}$ (12.16$^{**}$) & 3.17$^{**}$ (13.55$^{**}$) \\
Nash  \cite{navon2022multi} & \textcolor{second}{89.21} & 89.82 & \textcolor{second}{94.32} & \textcolor{second}{1.01} (5.51) & \textcolor{second}{0.42} (2.25) & 0.58 (3.39) & \textcolor{second}{68.40}$^{**}$ & \textcolor{second}{76.17}$^{**}$ & 82.35$^{**}$ & 6.19$^{**}$ (26.81$^{**}$) & 2.68$^{**}$ (12.96$^{**}$) & 3.45$^{**}$ (14.54$^{**}$) \\
AMTL  \cite{senushkin2023independent} & 89.02$^{**}$ & \textcolor{second}{89.85} & 94.24 & 1.15$^{**}$ (6.47$^{**}$) & \textcolor{second}{0.42} (\textcolor{best}{2.23}) & 0.58 (3.14$^{*}$) & 65.88$^{**}$ & 75.54$^{**}$ & 81.53$^{**}$ & 6.76$^{**}$ (27.97$^{**}$) & 2.52$^{**}$ (12.74$^{**}$) & 3.23$^{**}$ (13.58$^{**}$) \\
CAGrad  \cite{liu2021conflict} & 88.93$^{**}$ & 89.70 & 94.18 & 1.17$^{**}$ (6.63$^{**}$) & \textcolor{second}{0.42}$^{*}$ (2.25) & \textcolor{second}{0.54} (3.05) & 66.88$^{**}$ & 75.29$^{**}$ & \textcolor{second}{83.79}$^{**}$ & 6.52$^{**}$ (28.26$^{**}$) & 2.76$^{**}$ (13.76$^{**}$) & \textcolor{second}{2.78}$^{**}$ (\textcolor{second}{12.86}$^{**}$) \\
\midrule
ALBO (Ours) & \textcolor{best}{89.91} & \textcolor{best}{89.87} & \textcolor{best}{94.62} & \textcolor{best}{0.95} (\textcolor{second}{5.35}) & \textcolor{best}{0.40} (\textcolor{second}{2.23}) & \textcolor{best}{0.47} (\textcolor{best}{2.69}) & \textcolor{best}{70.74} & \textcolor{best}{80.21} & \textcolor{best}{87.48} & \textcolor{best}{4.29} (\textcolor{best}{16.03}) & \textcolor{best}{1.49} (\textcolor{best}{6.69}) & \textcolor{best}{2.19} (\textcolor{best}{9.94}) \\

\bottomrule
\end{tabular}
}

\end{table*}

\begin{table*}[t!]
\caption{Performance comparison on Hippocampus with 208 and 20 labeled scans for training.
Best and second-best values in each column are shown in \textcolor{best}{red} and \textcolor{second}{blue}, respectively.
Each ASSD (mm) value is followed by the 95\% Hausdorff Distance (HD95, mm) in parentheses. $^{*}$ and $^{**}$ indicate a statistically significant difference from ALBO with $p<0.05$ and $p<0.01$, respectively (paired Wilcoxon signed-rank test over all samples).
}\label{tab:hippocampus_effi}
\centering

\setlength{\tabcolsep}{6pt}
\resizebox{\textwidth}{!}{
\begin{tabular}{lcccccccc}
\toprule
\multirow{3}{*}{Method}&\multicolumn{4}{c}{\textbf{208 labeled scans}}&\multicolumn{4}{c}{\textbf{20 labeled scans}}\\
\cmidrule(lr){2-5}\cmidrule(lr){6-9}

&\multicolumn{2}{c}{Dice (\%) $\uparrow$}  & \multicolumn{2}{c}{ASSD (HD95) $\downarrow$}&\multicolumn{2}{c}{Dice (\%) $\uparrow$}  & \multicolumn{2}{c}{ASSD (HD95) $\downarrow$}  \\
			\cmidrule(lr){2-3}\cmidrule(lr){4-5}\cmidrule(lr){6-7}\cmidrule(lr){8-9}
		&	HH& HB   & HH & HB  &	HH& HB   & HH & HB \\
 \midrule
Independent & 88.56$^{**}$ & 86.87 & 0.39$^{**}$ (1.75$^{**}$) & 0.36$^{**}$ (1.54) & 85.02$^{**}$ & 82.81$^{**}$ & 0.66$^{**}$ (2.79$^{**}$) & 0.52$^{**}$ (2.20$^{**}$) \\
Joint & 88.89 & 86.95 & 0.35 (1.50) & 0.36 (1.71) & 85.63 & 83.31$^{**}$ & \textcolor{second}{0.51} (\textcolor{best}{2.03}) & 0.48$^{**}$ (1.94$^{**}$) \\
GCS  \cite{du2018adapting} & 88.74 & 86.99 & 0.36$^{**}$ (1.65$^{**}$) & 0.35 (1.53) & 85.76 & 83.74$^{**}$ & 0.53 (\textcolor{second}{2.07}) & 0.47$^{*}$ (1.94) \\
AuxiLearn  \cite{navon2020auxiliary} & \textcolor{second}{89.02} & 86.99 & 0.36 (1.58) & 0.36 (1.60) & 84.93$^{**}$ & 83.63$^{**}$ & 0.72$^{**}$ (3.94$^{**}$) & 0.48$^{**}$ (2.05$^{**}$) \\
AuxiNash  \cite{shamsian2023auxiliary} & \textcolor{best}{89.10} & \textcolor{best}{87.40}$^{**}$ & \textcolor{second}{0.34} (1.49$^{**}$) & \textcolor{best}{0.33}$^{**}$ (\textcolor{best}{1.44}) & 86.24$^{*}$ & 84.03 & 0.54$^{*}$ (2.22) & 0.45 (1.78) \\
RLW    \cite{lin2021reasonable} & 88.70$^{*}$ & \textcolor{second}{87.09} & 0.37$^{**}$ (1.73$^{**}$) & \textcolor{second}{0.34} (1.54) & \textcolor{second}{86.35} & 83.58$^{**}$ & \textcolor{best}{0.48} (2.10) & 0.47$^{**}$ (1.95$^{**}$) \\
Nash \cite{navon2022multi} & 88.96 & 86.75 & \textcolor{best}{0.33} (\textcolor{second}{1.39}) & 0.35 (\textcolor{second}{1.49}$^{*}$) & \textcolor{best}{86.40} & \textcolor{second}{84.05}$^{**}$ & \textcolor{second}{0.51} (2.27) & \textcolor{best}{0.43} (\textcolor{best}{1.71}) \\
AMTL \cite{senushkin2023independent} & 88.52$^{**}$ & 86.92 & 0.37$^{**}$ (1.59$^{*}$) & 0.35 (1.57) & 86.04 & 83.74$^{**}$ & 0.55 (2.49) & 0.46$^{*}$ (1.87$^{*}$) \\
CAGrad \cite{liu2021conflict} & 88.81 & 86.54$^{**}$ & 0.35$^{*}$ (1.64$^{**}$) & 0.37$^{**}$ (1.73) & 85.91 & 83.54$^{**}$ & 0.55 (2.64) & 0.47$^{**}$ (2.01$^{**}$) \\
\midrule
ALBO (Ours) & 89.01 & 87.08 & \textcolor{best}{0.33} (\textcolor{best}{1.38}) & \textcolor{best}{0.33} (1.54) & 86.25 & \textcolor{best}{84.51} & 0.52 (2.39) & \textcolor{second}{0.44} (\textcolor{second}{1.76}) \\

\bottomrule
\end{tabular}
}

\end{table*}

\subsubsection{Comparison on ACDC.}
For cardiac segmentation, besides using the full 160 scans in each fold, we also randomly selected 10, 20, 40, and 80 scans for training. 
In contrast to REFUGE, the models tended to \textit{underfit} on ACDC, and CCM provided more favorable results than BCM. Accordingly, we used U-Net-L for all compared methods to ensure a fair comparison.

Table \ref{tab:acdc_effi} presents the average performance on RV, Myo, and LV of all methods, trained with either 160 or 10 labeled scans.
ALBO consistently obtained the best results across all metrics, particularly when the labeled training data was limited to 10.
Table~\ref{tab:acdc_effi} also reports the HD95 and the significance of each method relative to ALBO (using a paired Wilcoxon signed-rank test). ALBO obtained the best Dice and ASSD on every structure and the best or second-best HD95; with only 10 labeled scans it significantly outperformed all competing methods on all three metrics for every structure ($p<0.01$).
Similarly, Fig.~\ref{fig:acdc_data_eff} (b) illustrates the trend of Dice of all methods averaged on all structures with varying numbers of labeled training data.
As the number of labeled scans decreases, the performance gap between ALBO and the competing methods widens, demonstrating the superior label efficiency of ALBO.
Specifically, ALBO significantly outperforms others in average Dice across all training sample sizes ($p<0.05$).
A similar trend is observed for ASSD.
The middle two rows of Fig.~\ref{fig:Visual_comparison} illustrate representative predictions from each method, trained with 160 and 10 scans, respectively. ALBO produces more accurate predictions for the sample, especially when labeled data were limited.

\subsubsection{Comparison on Hippocampus.} 
For hippocampus segmentation, we used the full set of 208 scans and randomly selected 20 scans for training, and employed U-Net-L for all compared methods.
Table \ref{tab:hippocampus_effi} presents the average results on the HH and HB.
Most methods outperformed the Independent when only 20 labeled scans were available, and ALBO remained robust in this low-label setting.
Table~\ref{tab:hippocampus_effi} additionally reports the HD95 and the per-metric significance relative to ALBO (using a paired Wilcoxon signed-rank test). Because this benchmark was close to saturation, all methods performed comparably and many pairwise differences were not statistically significant; ALBO nevertheless remained among the most competitive methods, achieving the best HD95 and the tied-best ASSD on the HH with 208 scans and the best HB Dice with 20 scans, while a few of the strongest auxiliary-learning methods such as AuxiNash and Nash were only marginally better on individual metrics.
The bottom two rows of Fig.~\ref{fig:Visual_comparison} illustrate representative predictions from each method, trained with 208 and 20 scans, respectively.
ALBO produces anatomically consistent segmentations.

\begin{table}[t]
\caption{Comparison with recent few-shot and semi-supervised segmentation methods on REFUGE with 1, 4 and 10 labeled images. \#Lab/\#Unl: number of labeled/unlabeled training images; \#Par: parameters (M); Time: per-image inference latency (ms) at the corresponding number of labeled images. ALBO and Independent use \emph{no} unlabeled data. $^{*}$ and $^{**}$ indicate a statistically significant difference from ALBO at $p<0.05$ and $p<0.01$ (paired Wilcoxon signed-rank test on per-image scores over three seeds). Best and second-best per column within each block are in \textcolor{best}{red} and \textcolor{second}{blue}. ``$/$'' denotes no meaningful ASSD is achieved overall;
these failures are reflected by the low Dice. 
}
\label{tab:refuge_fewshot}
\centering
\setlength{\tabcolsep}{2pt}
\resizebox{0.49\textwidth}{!}{
\begin{tabular}{lcccccccc}
\toprule
\multirow{2}{*}{Method} & \#Lab & \#Unl & \#Par & Time & \multicolumn{2}{c}{Dice (\%)$\uparrow$} & \multicolumn{2}{c}{ASSD (pixels)$\downarrow$} \\
\cmidrule(lr){6-7}\cmidrule(lr){8-9}
 & (img) & (img) & (M) & (ms) & OC & OD & OC & OD \\
\midrule
\multicolumn{9}{l}{\textbf{1 labeled image}} \\
\multicolumn{9}{l}{\quad\textit{Labeled-only (few-shot)}} \\
BRBS \cite{he2022brbs} & 1 & 0 & 3.93 & 7 & 34.61$^{**}$ & 50.31$^{**}$ & 68.01$^{**}$ & 79.98$^{**}$ \\
DSPNet \cite{tang2025dspnet} & 1 & 0 & 59.28 & 77 & 21.74$^{**}$ & 28.95$^{**}$ & / & / \\
UPRENet \cite{hu2025uprenet} & 1 & 0 & 51.60 & 781 & 45.47$^{**}$ & \textcolor{second}{69.29}$^{**}$ & / & \textcolor{second}{31.74}$^{**}$ \\
Independent & 1 & 0 & 0.50 & 14 & 39.81$^{**}$ & 62.44$^{**}$ & / & 59.95$^{**}$ \\
ALBO (Ours) & 1 & 0 & 0.50 & 14 & \textcolor{second}{60.42} & 66.90 & \textcolor{second}{24.44} & 54.28 \\
\addlinespace
\multicolumn{9}{l}{\quad\textit{Semi-supervised (uses unlabeled data)}} \\
BRBS \cite{he2022brbs} & 1 & 399 & 3.93 & 7 & \textcolor{best}{72.76}$^{**}$ & \textcolor{best}{77.25}$^{**}$ & \textcolor{best}{18.75}$^{**}$ & 44.66$^{**}$ \\
DSPNet \cite{tang2025dspnet} & 1 & 399 & 59.28 & 77 & 30.19$^{**}$ & 63.48$^{**}$ & / & \textcolor{best}{30.38}$^{**}$ \\
UPRENet \cite{hu2025uprenet} & 1 & 399 & 51.60 & 781 & 22.30$^{**}$ & 65.45$^{*}$ & / & 47.91$^{**}$ \\
\midrule
\multicolumn{9}{l}{\textbf{4 labeled images}} \\
\multicolumn{9}{l}{\quad\textit{Labeled-only (few-shot)}} \\
BRBS \cite{he2022brbs} & 4 & 0 & 3.93 & 7 & 41.86$^{**}$ & 52.80$^{**}$ & 68.85$^{**}$ & 85.75$^{**}$ \\
DSPNet \cite{tang2025dspnet} & 4 & 0 & 59.28 & 370 & 66.02$^{**}$ & 76.57$^{*}$ & 20.88$^{**}$ & \textcolor{second}{23.41}$^{**}$ \\
UPRENet \cite{hu2025uprenet} & 4 & 0 & 51.60 & 3132 & 66.61$^{**}$ & \textcolor{best}{85.22}$^{**}$ & 23.04$^{**}$ & \textcolor{best}{20.41}$^{**}$ \\
Independent & 4 & 0 & 0.50 & 14 & 56.29$^{**}$ & 74.68$^{**}$ & 21.82$^{**}$ & 42.39$^{**}$ \\
ALBO (Ours) & 4 & 0 & 0.50 & 14 & \textcolor{second}{68.96} & \textcolor{second}{77.99} & \textcolor{best}{18.24} & 37.90 \\
\addlinespace
\multicolumn{9}{l}{\quad\textit{Semi-supervised (uses unlabeled data)}} \\
BRBS \cite{he2022brbs} & 4 & 396 & 3.93 & 7 & \textcolor{best}{71.90}$^{**}$ & 74.49$^{**}$ & \textcolor{second}{20.09}$^{**}$ & 49.28$^{**}$ \\
DSPNet \cite{tang2025dspnet} & 4 & 396 & 59.28 & 370 & 54.84$^{**}$ & 72.57$^{**}$ & 26.06$^{**}$ & 26.43$^{**}$ \\
UPRENet \cite{hu2025uprenet} & 4 & 396 & 51.60 & 3132 & 40.69$^{**}$ & 67.70$^{**}$ & 40.62$^{**}$ & 48.57$^{**}$ \\
\midrule
\multicolumn{9}{l}{\textbf{10 labeled images}} \\
\multicolumn{9}{l}{\quad\textit{Labeled-only (few-shot)}} \\
BRBS \cite{he2022brbs} & 10 & 0 & 3.93 & 7 & 57.63$^{**}$ & 68.68$^{**}$ & 36.02$^{**}$ & 64.80$^{**}$ \\
DSPNet \cite{tang2025dspnet} & 10 & 0 & 59.28 & 632 & \textcolor{second}{69.00}$^{**}$ & \textcolor{second}{84.75}$^{**}$ & \textcolor{second}{19.96}$^{**}$ & \textcolor{best}{16.05}$^{**}$ \\
UPRENet \cite{hu2025uprenet} & 10 & 0 & 51.60 & 7850 & 68.89$^{**}$ & \textcolor{best}{85.64}$^{**}$ & 24.78$^{**}$ & \textcolor{second}{23.48}$^{**}$ \\
Independent & 10 & 0 & 0.50 & 14 & 67.54$^{**}$ & 80.91$^{**}$ & 22.83$^{**}$ & 30.50 \\
ALBO (Ours) & 10 & 0 & 0.50 & 14 & \textcolor{best}{73.58} & 82.42 & \textcolor{best}{18.76} & 28.32 \\
\addlinespace
\multicolumn{9}{l}{\quad\textit{Semi-supervised (uses unlabeled data)}} \\
BRBS \cite{he2022brbs} & 10 & 390 & 3.93 & 7 & 66.84$^{**}$ & 75.23$^{**}$ & 27.61$^{**}$ & 41.58$^{**}$ \\
DSPNet \cite{tang2025dspnet} & 10 & 390 & 59.28 & 632 & 67.78$^{**}$ & 81.10$^{**}$ & 24.64$^{**}$ & 25.82 \\
UPRENet \cite{hu2025uprenet} & 10 & 390 & 51.60 & 7850 & 57.22$^{**}$ & 78.39$^{**}$ & 35.20$^{**}$ & 37.49$^{**}$ \\
\bottomrule
\end{tabular}
}
\end{table}

\subsection{Discussion}

\subsubsection{Can ALBO learn task-consistent feature representations?}
In Fig. \ref{fig:refuge_feature}, we visualize encoder feature maps from the main and auxiliary branches for a representative retinal image using Independent, Joint and ALBO. 
Compared with Independent and Joint, ALBO produces feature representations that are more aligned with the corresponding tasks: the main-branch features are more focused on the segmentation target, while the auxiliary-branch features more clearly capture boundary-related structure. This suggests that ALBO encourages more task-consistent feature learning through bidirectional communication and bilevel optimization.

\subsubsection{How does ALBO compare with few-shot methods?}
\label{sec:fewshot_ssl}

ALBO is an auxiliary-learning method and does not require unlabeled data, which distinguishes it from semi-supervised learning (SSL). 
To position ALBO in the low-data setting, we compared it against three recent few-shot medical segmentation methods, including BRBS~\cite{he2022brbs}, DSPNet~\cite{tang2025dspnet} and UPRENet~\cite{hu2025uprenet}, on REFUGE with 1, 4 and 10 labeled images. 
Since these methods can also exploit unlabeled images, each was run in two modes: a labeled-only (few-shot) mode, and a semi-supervised mode that additionally used the remaining $390$--$399$ unlabeled training images. Table~\ref{tab:refuge_fewshot} reports accuracy together with model size and inference time.

The three baselines cover the two dominant paradigms of few-shot medical segmentation. BRBS is a \emph{registration-based} method following the registration-to-segmentation paradigm: it trains a deformable registration network on unlabeled images and generates a large pseudo-labeled training set by warping the few labeled atlases onto them, after which a compact segmentation network is trained on the synthetic set. It uses two small from-scratch U-Nets ($3.93$M parameters) and runs only the segmentation network at test time, so its inference is fast ($\sim$$7$\,ms/image) and independent of the label budget. DSPNet and UPRENet are \emph{prototypical} episodic methods with a COCO-pretrained ResNet-101 encoder, which explains their much larger parameter counts of $52$--$59$M. They embed the labeled support images, extract class prototypes, and segment each query by nearest-prototype matching; DSPNet refines pooling-based prototypes into detail-preserving cluster prototypes, whereas UPRENet additionally filters global and local prototypes by an uncertainty estimation and fuses two augmented support branches through an evidential module. Crucially, their $N$-shot setting requires extracting prototypes from \emph{every} labeled support image and averaging one prediction per support image at inference, so their latency grows roughly linearly with the number of labeled images, namely from $77$\,ms and $0.78$\,s at one label to $0.63$\,s and $7.8$\,s at ten labels for DSPNet and UPRENet, respectively. ALBO, BRBS and Independent instead run a single fixed network with constant latency ($\sim$$14$\,ms for ALBO), giving ALBO up to a near three-order-of-magnitude speed advantage at larger label budgets.

These architectural differences help explain the observed performance. BRBS is designed to exploit unlabeled images through registration-based label propagation. In the labeled-only mode, this mechanism is unavailable and the segmentation network is trained from only a few annotated images, resulting in substantially lower OC Dice ($34.6$--$57.6\%$). When unlabeled data are available,  BRBS performs best in the one-label setting ($72.8\%$), where a single annotated atlas can be propagated to many unlabeled images.
However, its performance \emph{degraded} as the number of labeled images increased. In contrast, the prototypical methods exhibit the opposite trend: with only one support image, a single prototype provides limited representation of the small, low-contrast, and highly variable optic cup, whereas performance improves as additional support prototypes become available.
Although these methods achieve good OD segmentation (up to $85.6\%$ Dice), they require $50$--$120\times$ more parameters and up to three orders of magnitude longer inference time. 
By comparison, ALBO attained the highest OC Dice among all labeled-only methods across all label budgets. With 10 labeled images it also surpassed every few-shot and SSL variant on OC segmentation, including those that consume hundreds of unlabeled images, while using only $0.50$M parameters and $\sim$$14$\,ms per image inference.

We emphasize that ALBO addresses a different problem from SSL: ALBO improves label efficiency through auxiliary labels derived from the existing annotations rather than through unlabeled data, making the two approaches complementary rather than competing. As shown in the following subsection, ALBO can be combined with SSL methods to achieve further improvements.

\begin{table}[t!]
\caption{Performance comparison for SSL methods on the ACDC dataset using 5-fold cross-validation with 10 labeled and 150 unlabeled scans for training in each fold.
Best and second-best values in each column are shown in \textcolor{best}{red} and \textcolor{second}{blue}, respectively.
}\label{tab:semi_acdc}
\centering
\setlength{\tabcolsep}{10pt}
\scalebox{1.0}{
\begin{tabular}{lcccc}
\toprule
\multirow{2}{*}{Method}&\multicolumn{4}{c}{Dice (\%) $\uparrow$} \\
			\cmidrule(lr){2-5}
		&	RV& Myo & LV  & Average  \\
 \midrule
 Independent  & 67.92   &  74.97  & 83.22 &75.37 \\
 \midrule
     \multicolumn{5}{l}{(a) SSL without auxiliary task (results cited from \cite{wu2022minimizing})}\\
 \midrule
PL  &  67.03 &  77.87 &  84.88 &76.59 \\

 AdvSemiSeg  & 73.11  & 80.18  &  86.47 &79.92 \\

 CPS  & 69.31  &  76.37 & 84.08 &76.59  \\
CrossTeaching  & 72.04  & 72.91  & 80.86 &75.27   \\
URPC & 60.29  & 66.30  &   74.67&67.09 \\
    MERU  &\textcolor{second}{74.90}   & \textcolor{best}{81.11}  &  \textcolor{best}{88.35} & \textcolor{second}{81.45} \\

  \midrule
    \multicolumn{5}{l}{(b) PL using boundary detection as the auxiliary task}\\
       \midrule
PL + Joint   &  63.77 & 77.04  &  84.56 & 75.12 \\
 PL + ALBO (Ours)   & 74.76  &  80.02 &  85.91  &80.23\\
     \midrule
    \multicolumn{5}{l}{(c) PL using distance-map prediction as the auxiliary task}\\
    \midrule
PL + Joint    &  69.23 &  78.69 &  86.02  & 77.98\\

 PL + ALBO (Ours)    &  \textcolor{best}{76.85} & \textcolor{second}{80.71}  & \textcolor{second}{87.50}  & \textcolor{best}{81.69} \\
\bottomrule
\end{tabular}
}

\end{table}

\begin{table*}[th!]
\caption{Performance comparison on the ACDC dataset using DINOv3 as the backbone with 4 and 10 labeled training scans. Best and second-best values in each column are shown in red and blue, respectively.
}\label{tab:dinov3_acdc_sm}
\centering
\setlength{\tabcolsep}{6pt}
\scalebox{1.0}{
\begin{tabular}{lcccccc|cccccc}
\toprule
\multirow{3}{*}{Method}&\multicolumn{6}{c|}{Auxiliary task: boundary detection} &\multicolumn{6}{c}{Auxiliary task:  distance-map prediction} \\
			\cmidrule(lr){2-7}\cmidrule(lr){8-13}
            & \multicolumn{3}{c}{Dice (\%) $\uparrow$}  & \multicolumn{3}{c|}{ASSD (mm) $\downarrow$}  & \multicolumn{3}{c}{Dice (\%) $\uparrow$}  & \multicolumn{3}{c}{ASSD (mm) $\downarrow$}\\
            \cmidrule(lr){2-4}\cmidrule(lr){5-7}\cmidrule(lr){8-10}\cmidrule(lr){11-13}
		&	RV& Myo & LV  & RV& Myo & LV  &	RV& Myo & LV  & RV& Myo & LV  \\
\midrule
 \multicolumn{13}{c}{Using \textit{4} labeled scans for training} \\
\midrule
 Independent  &  \textcolor{second}{75.77} & 76.94  &  \textcolor{second}{85.12} & \textcolor{second}{4.26} &2.10   &2.63 &  \textcolor{second}{75.77} & \textcolor{second}{76.94}  &  85.12 & 4.26 &2.10   &\textcolor{second}{2.63}  \\

 Joint  &  73.98 & \textcolor{second}{77.25}  & 83.99  & 5.08 & \textcolor{second}{2.03}  &\textcolor{second}{2.47} & 74.88 & 76.87  &\textcolor{second}{85.76} &\textcolor{second}{3.98}  & \textcolor{best}{1.79}  &\textcolor{best}{2.16} \\

 CAGrad \cite{liu2021conflict}  &  49.09 & 58.81  & 71.36  & 9.13 & 3.71  &5.15 & 42.11 & 53.04  & 67.10&  11.76& 5.44  &6.44 \\

 ALBO (Ours)  & \textcolor{best}{76.08}  &  \textcolor{best}{79.11} &\textcolor{best}{87.14}  & \textcolor{best}{3.82} &\textcolor{best}{1.50}  &\textcolor{best}{2.28} & \textcolor{best}{76.50} &  \textcolor{best}{78.03} &\textcolor{best}{86.79} &  \textcolor{best}{3.65}& \textcolor{second}{1.80}  &2.83 \\
\midrule
\multicolumn{13}{c}{Using \textit{10} labeled scans for training}\\
\midrule
 Independent  &  \textcolor{second}{83.00} & \textcolor{second}{82.53}  &  \textcolor{second}{89.70}  & \textcolor{second}{1.85} & 1.25  &\textcolor{best}{1.56} & 83.00 & \textcolor{best}{82.53}  &  \textcolor{best}{89.70}  & \textcolor{second}{1.85} & \textcolor{second}{1.25}  &\textcolor{best}{1.56}\\

 Joint  & 81.87  &  82.31 &  88.96  & 2.20 & \textcolor{second}{1.24}  &\textcolor{second}{1.75} &\textcolor{best}{83.92}  & 81.94  &\textcolor{second}{89.45} &1.86  &\textcolor{best}{1.20}   &\textcolor{best}{1.51} \\

 CAGrad \cite{liu2021conflict}  &   66.69  & 70.94  & 79.80 &5.15  & 2.55  &3.59 & 63.74 & 70.09  & 81.24& 4.85 & 2.29  & 2.94\\

 ALBO (Ours)  &  \textcolor{best}{83.25}  &  \textcolor{best}{83.08} & \textcolor{best}{90.28}   &\textcolor{best}{1.75}  & \textcolor{best}{1.23}  &1.83 & \textcolor{second}{83.59} &  \textcolor{second}{82.13} & 89.42& \textcolor{best}{1.74} &  1.41 & 1.95\\
\bottomrule
\end{tabular}
}
\end{table*}
\subsubsection{Can ALBO be used for semi-supervised learning (SSL)?}\label{sec:albo_ssl}
For SSL, the performance of models trained solely on labeled data often serves as a crucial foundation. 
This is particularly evident in methods that utilize pseudo-labeling, where the predictions on unlabeled data significantly influence model updates.
Given that ALBO demonstrates high label efficiency, it is expected to have the potential for SSL applications.

Generally, the objective of SSL can be formulated as $\mathcal{L}_{sup}(\mathcal{D}_l)+\lambda \mathcal{R}(\mathcal{D}_{ul})$, where $\mathcal{D}_l$ and $\mathcal{D}_{ul}$ denote the labeled and unlabeled data sets, $\mathcal{L}_{sup}$ represents the supervised loss, $\mathcal{R}$ denotes a regularization term on unlabeled data, and $\lambda$ is a balancing parameter.
Similar to Eq. \eqref{eq:ucb}, ALBO can be extended to SSL as follows,
\begin{subequations}
    \begin{align}
        (\mathrm{main}): \min_{\boldsymbol{\theta}_m,\boldsymbol{\theta}_a,\boldsymbol{\psi}_m} &\mathcal{L}_m(\boldsymbol{\theta}_m,\boldsymbol{\theta}_a,\boldsymbol{\psi}_m; \mathcal{D}_l) \notag\\
        &+ \lambda_m \mathcal{R}_m(\boldsymbol{\theta}_m,\boldsymbol{\theta}_a,\boldsymbol{\psi}_m; \mathcal{D}_{ul}) \label{eq:ucb_ssl:main}\\
        (\mathrm{auxiliary}):\quad  \min_{\boldsymbol{\theta}_a,\boldsymbol{\psi}_a}&\mathcal{L}_a(\boldsymbol{\theta}_m,\boldsymbol{\theta}_a,\boldsymbol{\psi}_a; \mathcal{D}_l) \notag\\
        &+ \lambda_a \mathcal{R}_a(\boldsymbol{\theta}_m,\boldsymbol{\theta}_a,\boldsymbol{\psi}_a; \mathcal{D}_{ul}).\label{eq:ucb_ssl:aux}
    \end{align}\label{eq:ucb_ssl}
\end{subequations}

Therefore, SSL methods that are capable of computing $\mathcal{R}_{m/a}$ can be  integrated with ALBO.
As a proof of concept, we use the standard Pseudo-label (PL) method \cite{lee2013pseudo} to validate this integration with ALBO on the ACDC dataset, denoted as PL + ALBO. 
We follow the evaluation protocol established in MERU \cite{wu2022minimizing}, and compare our approach with six SSL methods: PL  \cite{lee2013pseudo}, AdvSemiSeg \cite{arXiv-Adversarial-Hung}, CPS \cite{chen2021semi}, CrossTeaching \cite{luo2022semi}, URPC \cite{luo2022semi_media}, and MERU.

Table \ref{tab:semi_acdc}-(a) summarizes the performance of these SSL methods without the incorporation of any auxiliary tasks.
Among them, only AdvSemiSeg, CPS, and MERU consistently outperformed the baseline Independent across all three anatomical structures, whereas PL degrades performance on RV.
Table \ref{tab:semi_acdc}-(b) presents the results where boundary detection is used as the auxiliary task. In this setting, PL + ALBO improved performance, while the PL + Joint model was consistently worse than the original PL model in terms of Dice across all structures. Table \ref{tab:semi_acdc}-(c) shows the results when the distance-map prediction is used as the auxiliary task. Both PL + Joint and PL + ALBO performed better than when  boundary detection was used. Notably, PL + ALBO achieved the highest average Dice score of $81.69\%$ among all evaluated methods.
These results suggest that ALBO can complement SSL methods effectively. Future work may extend this analysis to additional SSL frameworks to further evaluate its generality and effectiveness.

\emph{Relationship between ALBO and SSL.} Both ALBO and SSL target the limited-annotation regime, but they rely on different resources. ALBO assumes only the existing annotations of the main task and generates its auxiliary labels from them without expanding the training set, whereas SSL improves label efficiency by additionally exploiting a pool of unlabeled images. This leads to three practical observations.
(i) \emph{Complementarity.} Because the two strategies draw on different sources of information, they are complementary rather than competing. As shown in Table~\ref{tab:semi_acdc}, combining ALBO with PL improved the average Dice of PL from $76.59\%$ to $80.23\%$ (boundary detection) and $81.69\%$ (distance-map prediction), whereas PL + Joint yielded only $75.12\%$ and $77.98\%$, respectively.
(ii) \emph{Exploiting unlabeled data.} ALBO itself does not require unlabeled data, but it can exploit them when they are available: as formulated in Eq.~\eqref{eq:ucb_ssl}, any SSL regularizer that can be computed on unlabeled images can be added to the main and auxiliary levels, while the bilevel structure that prioritizes the main task remains unchanged.
(iii) \emph{Practical scenarios.} ALBO is preferable when a sizeable and representative pool of unlabeled images is unavailable or cannot be used, for example, in small single-center cohorts or under data-sharing and privacy restrictions that limit the collection of additional images. Using only the 10 labeled ACDC scans, ALBO reached an average Dice of $79.48\%$ (Table~\ref{tab:acdc_effi}), which is higher than four of the six SSL methods in Table~\ref{tab:semi_acdc} that additionally used 150 unlabeled scans. When abundant unlabeled data from the same distribution are available, SSL is the natural choice, and ALBO can then serve as a stronger supervised component within it, as in PL + ALBO.

\subsubsection{Can ALBO be applied with pre-trained visual foundation models (FMs)?}
The preceding experiments evaluated the effectiveness of ALBO when using a convolutional U-Net trained from scratch.
Given the remarkable transferability of FMs, we next investigated whether auxiliary tasks can further enhance segmentation performance when ALBO is built on pre-trained models.
To this end, we use the recently released DINOv3 (ViT-B/16) model \cite{simeoni2025dinov3} as the backbone for evaluation.

As DINOv3 is a pre-trained ViT-based model, directly reformulating its transformer blocks into a BCM-style architecture may lead to catastrophic forgetting of the learned knowledge.
Instead, given $\mathbf{z}^l_{m}$ and $\mathbf{z}^l_{a}$, we generate $\mathbf{z}^{l+1}_{m}$ and $\mathbf{z}^{l+1}_{a}$ at the $l$-th ViT block ($l=1,...,12$) through a dual-path formulation:
\begin{subequations}
    \begin{align}
    &\mathit{step}\ 1:  \ \hat{\mathbf{z}}^l_{m}= \mathrm{LoRABlock}^{l}(\mathbf{z}^l_{m}), \ \hat{\mathbf{z}}^l_{a}= \mathrm{LoRABlock}^{l}(\mathbf{z}^l_{a}), \notag\\
&\mathit{step}\ 2: \ \mathbf{z}^{l+1}_{m} = \hat{\mathbf{z}}^l_{m} + \tau_m \hat{\mathbf{z}}^l_{a}, \ \mathbf{z}^{l+1}_{a} = \hat{\mathbf{z}}^l_{a} + \tau_a \hat{\mathbf{z}}^l_{m}, \notag
\end{align}
\end{subequations}
where LoRABlock$^l$ denotes the $l$-th transformer block wrapped with LoRA \cite{hu2022lora} for efficient fine-tuning, and $\tau_{m/a}$ are learnable scalars controlling feature communication via additive fusion.
Two BCM blocks are appended after the modified DINOv3 backbone to produce the final predictions.

We compared ALBO against Independent, Joint, and CAGrad, using both boundary detection and distance-map prediction as auxiliary tasks on the ACDC dataset with 4 and 10 labeled scans, respectively.
Following the standard LoRA fine-tuning protocol, all methods were optimized using the AdamW optimizer with an initial learning rate of 0.001 and evaluated under five-fold cross-validation.
The results are summarized in Table \ref{tab:dinov3_acdc_sm}.

With 10 labeled scans, ALBO achieved performance comparable to Independent, particularly when boundary detection was used as the auxiliary task.
Under the more challenging 4-label setting, ALBO outperformed both Independent and Joint, especially under the boundary detection setting.
In contrast, CAGrad exhibited a substantial performance degradation compared with Independent.

\begin{figure}[t]
\centering
\includegraphics[width=0.85\linewidth]{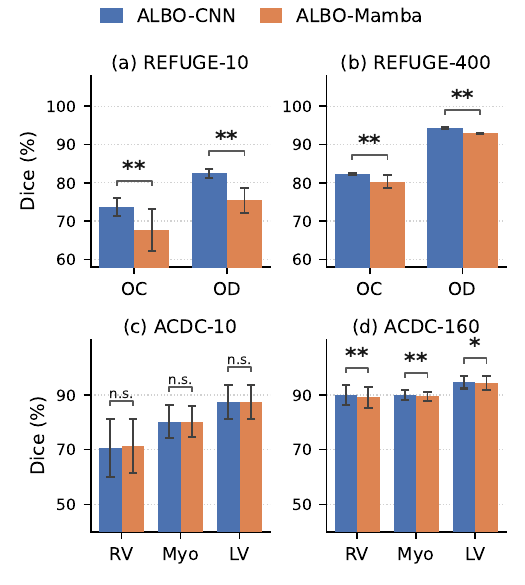}
\caption{ALBO with a convolutional (U-Net, blue) versus a Mamba (VSS, orange) backbone, using boundary detection as the auxiliary task. (a,b) REFUGE with 10 and 400 labeled images; (c,d) ACDC with 10 and 160 labeled scans. ``$*$'' and ``$**$'' denote $p<0.05$ and $p<0.01$ (paired Wilcoxon signed-rank test, Mamba vs.\ CNN); ``n.s.''\ = not significant. 
}
\label{fig:mamba}
\end{figure}

\subsubsection{Can ALBO be extended to the Mamba architecture?}
The preceding experiments evaluated ALBO on convolutional (U-Net) and transformer (DINOv3) backbones. We further examined whether ALBO extends to the Mamba~\cite{gu2023mamba} (state-space) architecture. To this end, we replaced the convolutional blocks in BCM with Vision-Mamba (VSS) blocks~\cite{zhu2024vim,liu2024vmamba}.

We compared the Mamba- and convolution-based ALBO using boundary detection as the auxiliary task, under both low and high label settings: REFUGE with 10 and 400 labeled images, and ACDC with 10 and 160 labeled scans. Fig.~\ref{fig:mamba} reports the per-structure Dice with paired Wilcoxon significance tests. On ACDC the Mamba backbone achieved performance comparable to the convolutional one. With 10 labeled scans, the two models were statistically indistinguishable on all structures (mean foreground Dice: $79.6\%$ vs.\ $79.5\%$), and with 160 labeled scans the difference remained below $1\%$. On REFUGE the Mamba backbone performed worse under the 10-label setting, but the gap shrank as the number of labels increased, from $6.5\%$ mean foreground Dice at 10 labels to $1.7\%$ at 400 labels.

These results indicate that ALBO generalizes naturally to Mamba-based architectures. 
The larger performance gap observed on REFUGE under extremely limited supervision is likely attributable to the backbone rather than the proposed ALBO, since state-space vision models are generally more data-hungry than convolutional networks and are not consistently superior across vision tasks. The key finding is that ALBO remains effective across convolutional, transformer, and state-space architectures.

\subsection{Computational and resource analysis}
We analyze the cost of the bilevel training scheme. Table~\ref{tab:cost} reports the training and inference cost of ALBO and of a single-level baseline (\textit{Independent}) on REFUGE with $N=10$ labeled images, measured on one NVIDIA RTX 6000 (24~GB) with batch size $20$.

\begin{table}[t]
\centering
\caption{Computational and resource overhead of ALBO versus the single-level baseline (\textit{Independent}) using the identical U-Net, on REFUGE ($N{=}10$).}
\label{tab:cost}
\footnotesize
\setlength{\tabcolsep}{4pt}
\begin{tabular}{lccc}
\toprule
Metric & Independent & ALBO & Ratio \\
\midrule
Parameters & 0.504~M & 0.504~M & $1.0\times$ \\
Training memory & 6.8~GB & 9.9~GB & $1.46\times$ \\
Time / iteration & 0.68~s & 3.81~s & $5.6\times$ \\
Iterations\ to best & 5040 & 2880 & $0.57\times$ \\
Wall-clock to best & 0.95~h & 3.05~h & $3.2\times$ \\
Inference latency & \multicolumn{2}{c}{$\approx$9--14~ms (same net)} & --- \\
Inference memory/params & \multicolumn{2}{c}{identical} & --- \\
\bottomrule
\end{tabular}
\end{table}

\begin{figure}[t]
\centering
\includegraphics[width=0.9\linewidth]{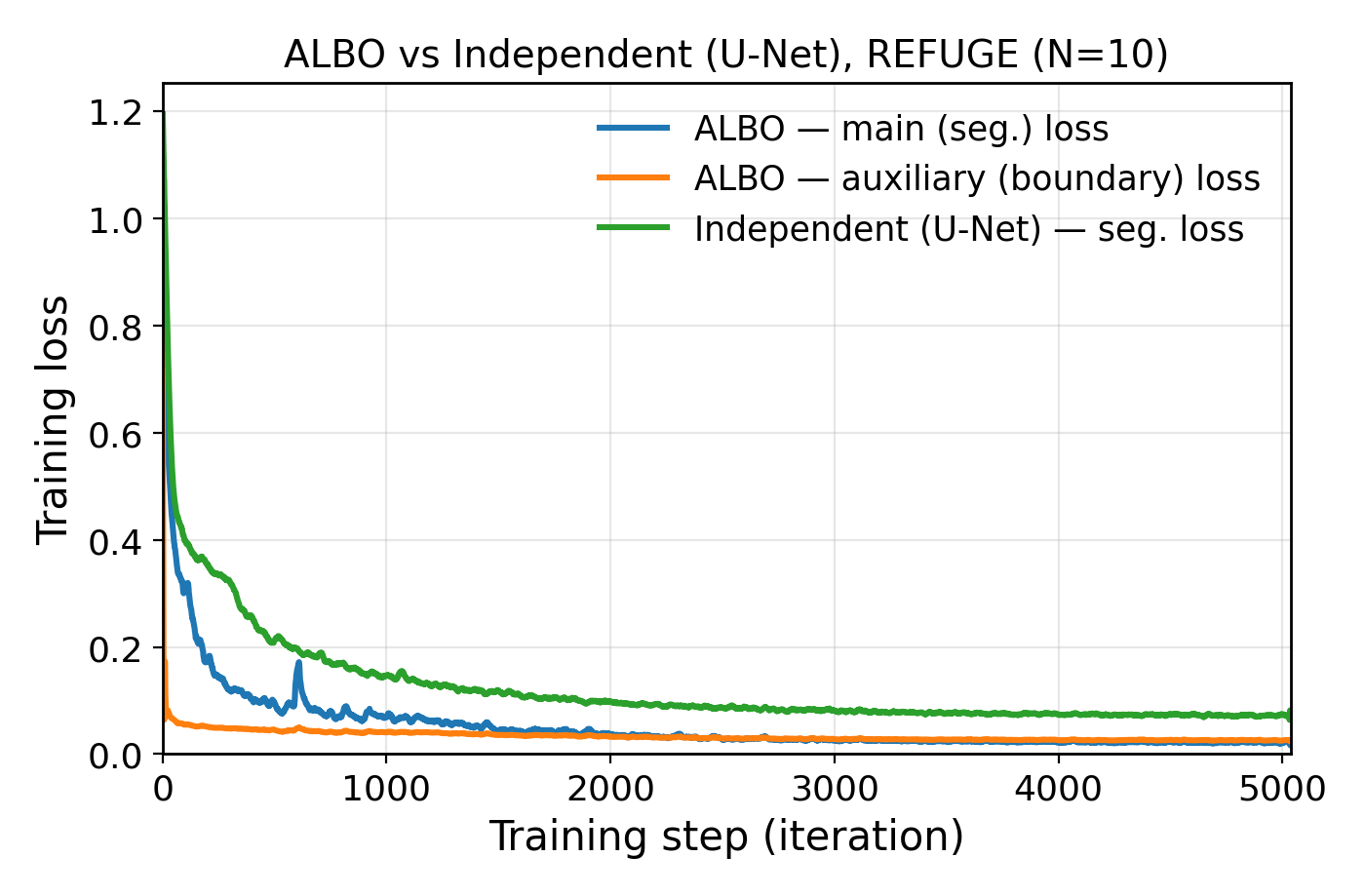}
\caption{Training loss convergence on REFUGE ($N{=}10$). ALBO's main-task (segmentation) loss decreases faster and to a lower value than the single-level baseline (\textit{Independent}).}
\label{fig:conv_loss}
\end{figure}

The additional computational cost arises from the bilevel optimization, which introduces one virtual inner update and a finite-difference approximation of the Hessian--vector product~\cite{liu2022auto}, requiring approximately two additional forward--backward passes per iteration. Consequently, the per-iteration training time increases by approximately $5.6\times$, and the peak training memory increases by 46\%. This overhead is incurred only during training. At inference, ALBO performs a single forward pass through the same segmentation network as the baseline, resulting in identical model size, memory footprint, and real-time inference latency ($<15$ ms per image). Furthermore, ALBO converges in fewer iterations (2880 versus 5040) while achieving a lower training loss (Fig.~\ref{fig:conv_loss}), reducing the wall-clock time to the best model from the expected $5.6\times$ to $3.2\times$. Overall, training completes in approximately three hours on a single GPU with a peak memory footprint below 10 GB.

\subsection{Failure cases and limitations}\label{sec:failure}
\begin{figure}[t]
\centering
\includegraphics[width=0.9\linewidth]{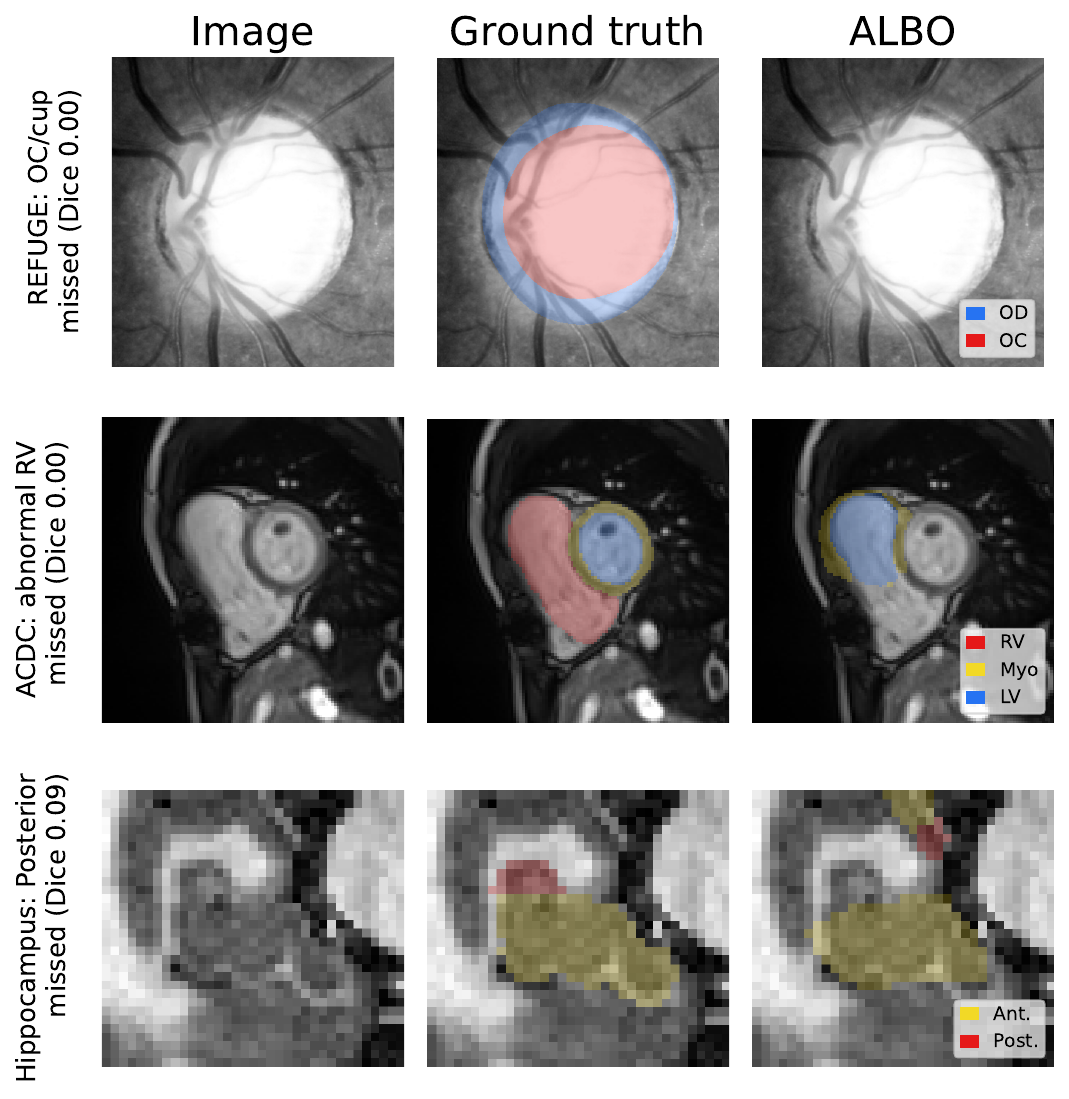}
\caption{Representative failure cases of ALBO under the 10-labeled setting on REFUGE (optic cup), ACDC (abnormal right ventricle), and Hippocampus (posterior). Each row shows the input image, ground truth, and ALBO prediction. The smallest, anatomically atypical, or lowest-contrast structure is frequently omitted and merged into the dominant neighboring structure.}
\label{fig:failure}
\end{figure}

Fig.~\ref{fig:failure} shows representative failure cases of ALBO under the 10-labeled setting on all three datasets. Most errors occur on the smallest, lowest-contrast, or anatomically atypical target, and ALBO tends to miss the target structure and assign its pixels to the neighboring structure (cup$\rightarrow$disc, posterior$\rightarrow$anterior, RV$\rightarrow$myocardium/background). On REFUGE, the OC is the dominant failure: its mean Dice ($73.58\%$) is well below the OD ($82.42\%$), and $10.4\%$ of test images have an OC Dice below $0.5$, including complete misses when the cup–disc boundary exhibits weak contrast. On ACDC, the RV is the most difficult structure and the most severe errors occur in the abnormal-RV pathology subgroup, where ten labeled scans are insufficient to capture the large anatomical variability. In addition, because the cardiac volumes are processed slice-by-slice using a 2D model, the lack of through-plane context further causes inter-slice inconsistencies on basal and apical slices. 
On the Hippocampus dataset, failures are concentrated in the posterior subdivision on thin head and tail slices, where the boundary between the anterior and posterior regions is ambiguous. In the most difficult cases, the posterior Dice decreases to $0.09$.

Despite its promising performance, ALBO has several limitations.
\begin{itemize}
    \item (i) \emph{Training cost.} The bilevel update requires a virtual inner step and a finite-difference Hessian--vector approximation, raising the per-iteration cost to $\approx\!5.6\times$ and the peak training memory by $46\%$ relative to single-level training (Table~\ref{tab:cost}). This cost is confined to the one-time, offline training stage, and inference is a single forward pass with identical model size and latency; nevertheless, it makes training more resource-intensive than conventional baselines.
    \item (ii) \emph{Limited gains with abundant labels.} The advantage of ALBO is most pronounced in the low-label regime; as the number of labeled samples grows, the gap to competing methods narrows. On near-saturated benchmarks (e.g., Hippocampus, and the $N{=}160/400$ settings of ACDC/REFUGE) ALBO remains competitive but is no longer uniformly best, and several differences do not reach statistical significance (Tables~\ref{tab:refuge_effi}--\ref{tab:hippocampus_effi}).
    \item (iii) \emph{Performance on rare structures.} Like other methods trained with very limited supervision, ALBO cannot span the full anatomical variability from ten labeled scans; it therefore has a ceiling on rare, atypical, or very small, low-contrast targets, such as the optic cup at $N{=}10$.
    \item (iv) \emph{Cross-domain generalization.} Our validation is conducted within each dataset; robustness under distribution shift across sites, scanners, or modalities is not evaluated here. As ALBO is a label-efficiency method that is orthogonal to domain-generalization and domain-adaptation techniques, integrating it with such techniques and assessing cross-domain robustness remain open.
    \item (v) \emph{Combination with other label-efficient methods.} ALBO improves label efficiency solely through auxiliary labels automatically derived from the labeled data, which is complementary to both few-shot and semi-supervised learning. 
  We show that ALBO is competitive with such methods (Sec.~\ref{sec:fewshot_ssl}) and can be paired with a semi-supervised objective (Sec.~\ref{sec:albo_ssl}); however, how best to unify auxiliary labels, few-shot priors, and unlabeled data within a single framework remains largely unexplored.
\end{itemize}
Future work will focus on improving robustness to small and anatomically atypical structures, extending ALBO to stronger 3D architectures and complementary auxiliary tasks, evaluating cross-domain generalization, and integrating ALBO more closely with few-shot and semi-supervised learning frameworks.

\begin{table*}[h]
\caption{ Performance of SOTA AL and MTL methods on the ACDC dataset using U-Net as the backbone and \textit{boundary detection} as the auxiliary task. We employed five-fold cross-validation, and used 160 labeled scans for training at each fold. U-Net uses 128 (or 1024) bottom channels with either the CCM or BCM module. Models tend to underfit with smaller networks, whereas ALBO is more robust than the compared methods when using the proposed BCM module.
Best and second-best values in each column are shown in \textcolor{best}{red} and \textcolor{second}{blue}, respectively.
}\label{tab:acdc_all_data}
\centering

\setlength{\tabcolsep}{8pt}
\begin{tabular}{lcccccccc}
\toprule
			\multirow{2}{*}{Methods}&\multicolumn{4}{c}{Dice (\%) $\uparrow$}& \multicolumn{4}{c}{ASSD (mm) $\downarrow$} \\
			\cmidrule(lr){2-5}\cmidrule(lr){6-9}
		&	RV& Myo &LV & Avg   & RV& Myo &LV & Avg   \\
\midrule

 \multicolumn{9}{c}{(a) \textit{CCM}, U-Net-L (1024 ch., $\approx$31.05\,M params)} \\
\midrule
  Independent     & 89.16   &  89.47 &  94.01  &  90.88  & 1.03   &  0.44 &  0.63  &  0.71   \\

 Joint   & 88.93   &  89.56 &  94.19  &  90.89  & 1.11   &  0.45 &  0.63  &  0.73   \\

 GCS \cite{du2018adapting}      & \textcolor{second}{89.20}   &  89.49 &  94.14  &  90.94  & 1.04   &  0.46 &  0.61  &  0.71   \\

 AuxiLearn \cite{navon2020auxiliary} &89.09 &89.47 &94.15 & 90.91 &\textcolor{second}{1.03} &0.45 &0.56 &\textcolor{second}{0.68} \\

AuxiNash \cite{shamsian2023auxiliary} &89.06 &\textcolor{best}{89.85} &\textcolor{second}{94.31} & \textcolor{second}{91.07} &1.14 &\textcolor{best}{0.42} &\textcolor{second}{0.54} &0.70 \\

 RLW \cite{lin2021reasonable}    & 89.18   &  89.43 &  94.22  &  90.94  & 1.04   &  0.45 &  0.59  &  0.70   \\

 Nash \cite{navon2022multi}     & \textcolor{best}{89.21}   &  89.82 &  \textcolor{best}{94.32}  &  \textcolor{best}{91.12}  & \textcolor{best}{1.01}   &  \textcolor{best}{0.42} &  0.58  &  \textcolor{best}{0.67}   \\

AMTL \cite{senushkin2023independent}      & 89.02   &  \textcolor{second}{89.85} &  94.24  &  91.04  & 1.15   &  \textcolor{best}{0.42} &  0.58  &  0.72   \\

 CAGrad \cite{liu2021conflict}     & 88.93   &  89.70 &  94.18  &  90.94  & 1.17   &  \textcolor{second}{0.42} &  \textcolor{best}{0.54}  &  0.71   \\
\midrule
 \multicolumn{9}{c}{(b) \textit{BCM}, U-Net-L (1024 ch., $\approx$32.04\,M params)} \\
\midrule

  Independent     & 89.49   &  89.68 &  94.44  &  91.20  & 0.99   &  0.43 &  0.58  &  0.67   \\

 Joint   & 89.62   &  89.94 &  94.52  &  91.36  & 1.04   &  0.42 &  0.53  &  0.67   \\

 GCS \cite{du2018adapting}    & \textcolor{second}{89.87}   &  89.89 &  \textcolor{second}{94.70}  &  91.49  & \textcolor{second}{0.96}   &  0.42 &  0.48  &  0.62   \\

AuxiLearn \cite{navon2020auxiliary} &89.53 &89.82 &94.43 &  91.26&1.01 &0.42 &0.52 &0.65 \\

AuxiNash \cite{shamsian2023auxiliary} &89.80 &\textcolor{second}{90.09} &94.64 &\textcolor{second}{91.51}  &1.11 &0.41 &0.52 & 0.68\\

 RLW \cite{lin2021reasonable}      & 89.56   &  89.83 &  94.35  &  91.25  & 0.97   &  0.44 &  0.60  &  0.67   \\

 Nash \cite{navon2022multi}     & 89.62   &  89.98 &  94.48  &  91.36  & 1.01   &  0.42 &  0.54  &  0.66   \\

AMTL \cite{senushkin2023independent}      & 89.41   &  89.88 &  94.50  &  91.26  & 1.05   &  0.43 &  0.59  &  0.69   \\

 CAGrad \cite{liu2021conflict}       & 89.70   &  \textcolor{best}{90.18} &  \textcolor{best}{94.87}  &  \textcolor{best}{91.58}  & 0.99   &  \textcolor{best}{0.38} &  \textcolor{best}{0.44}  &  \textcolor{second}{0.61}   \\

 ALBO (Ours)   & \textcolor{best}{89.91}   &  89.87 &  94.62 &  91.47  & \textcolor{best}{0.95}   &  \textcolor{second}{0.40} &  \textcolor{second}{0.47}  &  \textcolor{best}{0.60}   \\
\midrule

 \multicolumn{9}{c}{(c) \textit{CCM}, U-Net-S (128 ch., $\approx$0.49\,M params)} \\
\midrule
  Independent     & 87.15   &  87.31 &  93.08  &  89.18  & \textcolor{best}{1.30}   &  0.56 &  0.72  &  \textcolor{best}{0.86}   \\

 Joint   & \textcolor{best}{87.36}   &  87.34 &  \textcolor{second}{93.28}  &  \textcolor{second}{89.33}  & \textcolor{second}{1.32}   &  \textcolor{second}{0.56} &  0.73  &  \textcolor{second}{0.87}   \\

 GCS \cite{du2018adapting}      & 86.75   &  87.09 &  92.99  &  88.94  & 1.46   &  0.59 &  0.72  &  0.93   \\

AuxiLearn \cite{navon2020auxiliary} &86.96 &86.98 &92.82 & 88.92 &1.38 &0.60 &0.77 & 0.91\\

AuxiNash \cite{shamsian2023auxiliary} &\textcolor{second}{87.26} &\textcolor{best}{87.84} &\textcolor{best}{93.55} &\textcolor{best}{89.55}  &1.35 &0.57 &0.71 &0.88 \\

 RLW \cite{lin2021reasonable}    & 86.95   &  87.09 &  93.12  &  89.05  & 1.47   &  0.57 &  \textcolor{best}{0.69}  &  0.91   \\

 Nash \cite{navon2022multi}     & 86.82   &  \textcolor{second}{87.67} & 93.12  &  89.20  & 1.46   &   \textcolor{best}{0.55} &  \textcolor{second}{0.70}  &  0.90   \\

AMTL \cite{senushkin2023independent}      & 85.76   &  87.29 &  92.58  &  88.54  & 1.80   &  0.60 &  0.96  &  1.12   \\

 CAGrad \cite{liu2021conflict}     & 86.75   &  87.26 &  93.23  &  89.08  & 1.47   &  0.58 &  0.79  &  0.95   \\
\midrule

 \multicolumn{9}{c}{(d) \textit{BCM}, U-Net-S (128 ch., $\approx$0.50\,M params)} \\
\midrule
   Independent    & 85.52   &  84.58 &  91.25  &  87.12  & 1.61   &  0.72 &  0.87  &  1.06   \\

 Joint   & 85.77   &  85.47 &  91.51  &  86.77  & 1.70   &  0.72 &  0.94  &  1.22   \\

 GCS \cite{du2018adapting}      & 84.02   &  85.49 &  91.82  &  87.11  & 1.99   &  0.65 &  0.89  &  1.18   \\

AuxiLearn \cite{navon2020auxiliary} &86.10 &85.52 &92.22 & 87.94 &\textcolor{second}{1.44} &0.70 &0.90 & 1.01\\

AuxiNash \cite{shamsian2023auxiliary} &\textcolor{second}{86.61} &\textcolor{second}{86.83} &\textcolor{second}{93.04} & \textcolor{second}{88.83} &1.58 &\textcolor{second}{0.60} &\textcolor{second}{0.75} & \textcolor{second}{0.98}\\

 RLW \cite{lin2021reasonable}    & 84.93   &  84.83 &  90.65  &  86.80  & 1.82   &  0.73 &  0.99  &  1.18   \\

 Nash \cite{navon2022multi}     & 85.42   &  86.24 &  92.07  &  87.91  & 1.76   &  0.70 &  0.90  &  1.12   \\

AMTL \cite{senushkin2023independent}      & 86.08   &  86.02 &  91.81  &  87.97  & 1.57   &  0.67 &  0.86  &  1.03   \\

 CAGrad \cite{liu2021conflict}     & 85.42   &  85.82 &  91.29  &  87.51  & 1.65   &  0.69 &  0.97  &  1.10   \\

ALBO (Ours)     & \textcolor{best}{88.26}   &  \textcolor{best}{88.49} &  \textcolor{best}{93.91}  &  \textcolor{best}{90.22}  & \textcolor{best}{1.18}   &  \textcolor{best}{0.49} &  \textcolor{best}{0.56}  &  \textcolor{best}{0.74}   \\
\bottomrule
\end{tabular}

\end{table*}

\begin{table}[!h]
\caption{ Performance of compared methods on the REFUGE 2018 dataset using U-Net as the backbone with 128 (or 1024) bottom channels and CCM (or BCM) modules. We conducted all experiments three times with different random seeds, and used 400 labeled images for training at each run. We adopted \textit{boundary detection} as the auxiliary task. 
The results show that models tend to overfit with larger networks, while our model performed more robustly than others. 
Additionally, other models showed better performance when using the proposed BCM module. \#Params denotes the parameter count of the model. 
Best and second-best values in each column are shown in \textcolor{best}{red} and \textcolor{second}{blue}, respectively.
}\label{tab:refuge_all_data}
\centering

\setlength{\tabcolsep}{12pt}
\begin{tabular}{lcccc}
\toprule
			\multirow{2}{*}{Methods}&\multicolumn{2}{c}{Dice (\%) $\uparrow$}  & \multicolumn{2}{c}{ASSD (pixels) $\downarrow$} \\
			\cmidrule(lr){2-3}\cmidrule(lr){4-5}
		&	OC& OD   & OC & OD   \\
\midrule
 \multicolumn{5}{c}{(a) \textit{CCM}, U-Net-S (128 ch., $\approx$0.49\,M params)} \\
 \midrule
  Independent    &  74.77 &  91.18  & 21.94   &  14.51  \\

 Joint &  74.65 &  90.93  & 22.08   &  13.79  \\

 GCS \cite{du2018adapting}    &  76.97 &  92.16  & 18.58   &  11.37  \\

AuxiLearn \cite{navon2020auxiliary} &75.65 &88.94 &18.72 &17.87  \\

AuxiNash \cite{shamsian2023auxiliary} &72.58 &86.40 &21.18 &22.93  \\

 RLW \cite{lin2021reasonable}   &  \textcolor{second}{78.02} &  \textcolor{best}{92.79}  & \textcolor{second}{18.24}   &  \textcolor{second}{10.64}  \\

 Nash \cite{navon2022multi}    &  76.10 &  91.42  & 19.05   &  12.25  \\

AMTL \cite{senushkin2023independent}     &  73.53 &  90.37  & 22.47   &  15.04  \\

 CAGrad \cite{liu2021conflict}   &  \textcolor{best}{78.07} &  \textcolor{second}{92.60}  & \textcolor{best}{17.99}   &  \textcolor{best}{10.56}  \\
\midrule

 \multicolumn{5}{c}{(b) \textit{BCM}, U-Net-S (128 ch., $\approx$0.50\,M params)} \\
\midrule
  Independent    &  81.54 &  93.17  & 10.28   &  \ 7.74  \\

 Joint &  \textcolor{second}{82.91} &  93.09  & \ 9.62   &  \ 7.86  \\

 GCS \cite{du2018adapting}    &  \textcolor{best}{83.39} &  \textcolor{second}{93.78}  & \textcolor{best}{\ 9.30}   &  \textcolor{second}{\ 6.98}  \\

AuxiLearn \cite{navon2020auxiliary} &82.75 &93.32 &\ 9.93 &7.53  \\

AuxiNash \cite{shamsian2023auxiliary} &81.37 &93.07 &10.36 &7.81  \\

 RLW \cite{lin2021reasonable}   &  81.05 &  93.42  & 10.31   &  \ 7.44  \\

 Nash \cite{navon2022multi}    &  80.97 &  93.11  & 10.76   &  \ 7.84  \\

AMTL \cite{senushkin2023independent}     &  81.16 &  93.66  & 10.49   &  \ 7.07  \\

 CAGrad \cite{liu2021conflict}   &  82.55 &  93.62  & \ 9.60   &  \ 7.15  \\

 ALBO (Ours)    & 82.30   &  \textcolor{best}{94.32} & \textcolor{second}{\ 9.36}   &  \textcolor{best}{\ 6.38}  \\
\midrule
 \multicolumn{5}{c}{(c) \textit{CCM}, U-Net-L (1024 ch., $\approx$31.05\,M params)} \\
 \midrule
  Independent    &  71.98 &  \textcolor{second}{84.87}  & 25.30   &  25.14  \\

 Joint &  69.91 &  82.68  & 30.38   &  28.95  \\

 GCS \cite{du2018adapting}    &  71.89 &  \textcolor{best}{86.47}  & 23.98   &   \textcolor{best}{22.71}  \\

AuxiLearn \cite{navon2020auxiliary} &\ 49.43 &60.13 &60.43 &70.48  \\

AuxiNash \cite{shamsian2023auxiliary} &44.60 &60.52 &70.18 &68.21  \\

 RLW \cite{lin2021reasonable}   &  \textcolor{second}{72.65} &  82.35  & \textcolor{second}{21.75}   &  28.08  \\

 Nash \cite{navon2022multi}    &  \textcolor{best}{74.27} &  84.69  & \textcolor{best}{19.05}   &  \textcolor{second}{24.61}  \\

AMTL \cite{senushkin2023independent}     &  71.92 &  83.97  & 28.92   &  25.47  \\

 CAGrad \cite{liu2021conflict}   &  71.05 &  83.86  & 24.13   &  25.66  \\
\midrule

 \multicolumn{5}{c}{(d) \textit{BCM}, U-Net-L (1024 ch., $\approx$32.04\,M params)} \\
\midrule
  Independent    &  \textcolor{second}{77.92} &  \textcolor{second}{90.82}  & \textcolor{second}{13.89}   &  \textcolor{second}{11.41}  \\

 Joint &  71.43 &  89.73  & 20.32   &  12.66  \\

 GCS \cite{du2018adapting}    &  75.78 &  90.63  & 15.33   &  12.76  \\

AuxiLearn \cite{navon2020auxiliary} &44.59 &65.12 &64.02 &57.98  \\

AuxiNash \cite{shamsian2023auxiliary} &44.97 &58.49 &66.91 &74.29  \\

 RLW \cite{lin2021reasonable}   &  74.43 &  88.74  & 15.82   &  14.21  \\

 Nash \cite{navon2022multi}    &  75.05 &  88.50  & 16.67   &  15.10  \\

AMTL \cite{senushkin2023independent}     &  68.30 &  85.95  & 24.66   &  19.49  \\

 CAGrad \cite{liu2021conflict}   &  72.43 &  87.53  & 21.43   &  16.85  \\

 ALBO (Ours) &  \textcolor{best}{80.47} &  \textcolor{best}{93.60}  & \textcolor{best}{11.45}   &  \textcolor{best}{\ 7.53}  \\
\bottomrule
\end{tabular}
\end{table}

\section{Conclusion}
In this work, we investigated AL for medical image segmentation, focusing on auxiliary tasks whose labels can be automatically derived from existing segmentation annotations without additional manual labeling. We proposed ALBO, a bilevel auxiliary learning framework that combines bidirectional feature communication with bilevel optimization to improve the primary segmentation task. Extensive experiments on three medical imaging benchmarks demonstrated that ALBO consistently improves label efficiency, particularly in low-label settings, while generalizing across convolutional, transformer, and state-space backbones. These results suggest that automatically generated auxiliary tasks, when coupled with appropriate architectural and optimization strategies, provide an effective and practical approach to label-efficient medical image segmentation.

\appendices

\section{Derivation of the finite-difference approximation}\label{app:fd}

We provide the full derivation of the look-ahead gradient summarized in Eq.~\eqref{eq:fid}. Recall the inner (auxiliary) update $\boldsymbol{\theta}_a^{'}=\boldsymbol{\theta}_a-\alpha\triangledown_{\boldsymbol{\theta}_a}\mathcal{L}_a(\boldsymbol{\theta}_m,\boldsymbol{\theta}_a,\boldsymbol{\psi}_a)$. Applying the chain rule through $\boldsymbol{\theta}_a^{'}$ and approximating the resulting Hessian--vector products by central finite differences yields, for the main branch,
    \begin{align}
        &\triangledown_{\boldsymbol{\theta}_m}\mathcal{L}_m(\boldsymbol{\theta}_m,\boldsymbol{\theta}_a^{'},\boldsymbol{\psi}_m)\nonumber\\
        \approx& \triangledown_{\boldsymbol{\theta}_m}\mathcal{L}_m(\boldsymbol{\theta}_m,\boldsymbol{\theta}_a - \alpha\triangledown_{\boldsymbol{\theta}_a}\mathcal{L}_a(\boldsymbol{\theta}_m,\boldsymbol{\theta}_a,\boldsymbol{\psi}_a),\boldsymbol{\psi}_m) \nonumber\\
        =&\triangledown_{\boldsymbol{\theta}_m}\mathcal{L}_m(\boldsymbol{\theta}_m,\boldsymbol{\theta}_a^{'},\boldsymbol{\psi}_m)- \nonumber\\
   &\triangledown_{\boldsymbol{\theta}_a^{'}}\mathcal{L}_m(\boldsymbol{\theta}_m,\boldsymbol{\theta}_a^{'},\boldsymbol{\psi}_m)\cdot \alpha \triangledown^2_{\boldsymbol{\theta}_m,\boldsymbol{\theta}_a}\mathcal{L}_a(\boldsymbol{\theta}_m,\boldsymbol{\theta}_a,\boldsymbol{\psi}_a) \nonumber\\
        \approx& \triangledown_{\boldsymbol{\theta}_m}\mathcal{L}_m(\boldsymbol{\theta}_m,\boldsymbol{\theta}_a^{'},\boldsymbol{\psi}_m)- \nonumber\\
          &\alpha \frac{\triangledown_{\boldsymbol{\theta}_m}\mathcal{L}_a(\boldsymbol{\theta}_m,\boldsymbol{\theta}_a^{+},\boldsymbol{\psi}_a)-\triangledown_{\boldsymbol{\theta}_m}\mathcal{L}_a(\boldsymbol{\theta}_m,\boldsymbol{\theta}_a^{-},\boldsymbol{\psi}_a)}{2\epsilon}, \label{eq:fid_m}
    \end{align}
and, for the auxiliary branch,
\
    \begin{align}
        &\triangledown_{\boldsymbol{\theta}_a}\mathcal{L}_m(\boldsymbol{\theta}_m,\boldsymbol{\theta}_a^{'},\boldsymbol{\psi}_m) \nonumber\\
        \approx& \triangledown_{\boldsymbol{\theta}_a}\mathcal{L}_m(\boldsymbol{\theta}_m,\boldsymbol{\theta}_a - \alpha\triangledown_{\boldsymbol{\theta}_a}\mathcal{L}_a(\boldsymbol{\theta}_m,\boldsymbol{\theta}_a,\boldsymbol{\psi}_a),\boldsymbol{\psi}_m) \nonumber\\
        =&\triangledown_{\boldsymbol{\theta}_a^{'}}\mathcal{L}_m(\boldsymbol{\theta}_m,\boldsymbol{\theta}_a^{'},\boldsymbol{\psi}_m)- \nonumber\\
        &\triangledown_{\boldsymbol{\theta}_a^{'}}\mathcal{L}_m(\boldsymbol{\theta}_m,\boldsymbol{\theta}_a^{'},\boldsymbol{\psi}_m)\cdot \alpha \triangledown^2_{\boldsymbol{\theta}^2_a}\mathcal{L}_a(\boldsymbol{\theta}_m,\boldsymbol{\theta}_a,\boldsymbol{\psi}_a) \nonumber\\
        \approx &\triangledown_{\boldsymbol{\theta}_a^{'}}\mathcal{L}_m(\boldsymbol{\theta}_m,\boldsymbol{\theta}_a^{'},\boldsymbol{\psi}_m) -  \nonumber\\
        &\alpha \frac{\triangledown_{\boldsymbol{\theta}_a^{+}}\mathcal{L}_a(\boldsymbol{\theta}_m,\boldsymbol{\theta}_a^{+},\boldsymbol{\psi}_a)-\triangledown_{\boldsymbol{\theta}_a^{-}}\mathcal{L}_a(\boldsymbol{\theta}_m,\boldsymbol{\theta}_a^{-},\boldsymbol{\psi}_a)}{2\epsilon},\label{eq:fid_a}
    \end{align}
where $\boldsymbol{\theta}_a^{\pm}=\boldsymbol{\theta}_a \pm \epsilon\cdot\triangledown_{\boldsymbol{\theta}_a^{'}}\mathcal{L}_m(\boldsymbol{\theta}_m,\boldsymbol{\theta}_a^{'},\boldsymbol{\psi}_m)$ and $\epsilon$ is a small positive value. Writing the two cases compactly with the shorthand $\boldsymbol{\theta}_{m/a}$ recovers Eq.~\eqref{eq:fid} in the main text.

\section{Effect of the BCM and CCM Modules across Network Scales}\label{app:bcm}
To support the discussion of the bidirectional communication module (Sec.~\ref{method:CCB}) and the bilevel optimization (Sec.~\ref{method:PF}), this appendix reports the full comparison between the BCM and CCM modules under two network scales, U-Net-S ($128$ bottom channels) and U-Net-L ($1024$ bottom channels), on the ACDC and REFUGE datasets. For ACDC, we used five-fold cross-validation with $160$ labeled scans per fold. 
For REFUGE, each model was trained three times with different random initializations using $400$ labeled images. 
Tables~\ref{tab:acdc_all_data}--\ref{tab:refuge_all_data} summarize the results.

On ACDC, the models are prone to underfitting (Table~\ref{tab:acdc_all_data}). With U-Net-L, CCM and BCM give comparable results for all methods. With the smaller U-Net-S, CCM is preferable to BCM for the nine compared methods; nevertheless, ALBO with BCM still attains the best average Dice ($90.22\%$). 

On REFUGE, embedding the BCM consistently helps at the small scale: with U-Net-S, all nine compared methods improve significantly over CCM (paired $t$-test, $p \ll 0.01$; Table~\ref{tab:refuge_all_data}, panels (a)--(b)), opposite to the trend on ACDC. When the network is enlarged to U-Net-L, all nine methods degrade markedly due to overfitting, under both CCM and BCM, whereas ALBO remains the most robust (panels (c)--(d)).

These observations indicate that the benefit of the BCM is not automatic: it depends on the dataset and the network scale, and the bilevel optimization is what enables the BCM to be exploited robustly across settings.

\bibliographystyle{IEEEtran}

\bibliography{refere}

\end{document}